\PassOptionsToPackage{prologue,dvipsnames,table}{xcolor}
\DocumentMetadata{}
\documentclass[sigconf]{acmart}

\AtBeginDocument{%
  \providecommand\BibTeX{{%
    \normalfont B\kern-0.5em{\scshape i\kern-0.25em b}\kern-0.8em\TeX}}}

\setcopyright{none}
\usepackage{comment}
\usepackage{enumitem}
\usepackage{booktabs}
\usepackage{soul}
\usepackage{amsmath}
\usepackage[table]{xcolor}
\usepackage{tcolorbox}
\usepackage[ruled, vlined, linesnumbered]{algorithm2e}
\usepackage{subfigure}
\usepackage{multirow}
\usepackage{adjustbox}
\usepackage{bm}
\usepackage{pgfplots}
\usetikzlibrary{matrix}
\usetikzlibrary{patterns}
\usepgfplotslibrary{groupplots}
\pgfplotsset{compat=newest}
\usepgfplotslibrary{external}
\usepgfplotslibrary{statistics}

\begin{document}

\title{Uncertain Multi-Objective Recommendation via Orthogonal Meta-Learning Enhanced Bayesian Optimization}

\renewcommand{\shorttitle}{Uncertain Multi-Objective Recommendation}

\author{Hongxu Wang}
\authornote{Both authors contributed equally to this research.}
\affiliation{
\institution{Chongqing University}
\city{Chongqing}
\country{China}
}
\email{hxwang0721@outlook.com}
\author{Zhu Sun}
\authornotemark[1]
\affiliation{%
  \institution{Singapore University of Technology and Design}
  \city{}
  \country{Singapore}
}
\email{sunzhuntu@gmail.com}

\author{Yingpeng Du}
\authornote{Corresponding author.}
\affiliation{%
  \institution{Nanyang Technological University}
  \city{}
  \country{Singapore}
}
\email{dyp1993@pku.edu.cn}

\author{Lu Zhang}
\affiliation{%
  \institution{Chengdu University of Information Technology}
  \city{Chengdu}
  \country{China}
}
\email{zhang\_lu010@outlook.com}

\author{Tiantian He}
\affiliation{%
 \institution{Agency for Science, Technology and Research}
 \city{}
 \country{Singapore}
}
\email{he\_tiantian@cfar.a-star.edu.sg}

\author{Yew-Soon Ong}
\affiliation{%
  \institution{Nanyang Technological University}
  \city{}
  \country{Singapore}}
\email{asysong@ntu.edu.sg}



\begin{abstract}

Recommender systems (RSs) play a crucial role in shaping our digital interactions, influencing how we access and engage with information across various domains. Traditional research has predominantly centered on maximizing recommendation accuracy, often leading to unintended side effects such as echo chambers and constrained user experiences. Drawing inspiration from autonomous driving, we introduce a novel framework that categorizes RS autonomy into five distinct levels, ranging from basic rule-based accuracy-driven systems to behavior-aware, uncertain multi-objective RSs—where users may have varying needs, such as accuracy, diversity, and fairness. In response, we propose an approach that dynamically identifies and optimizes multiple objectives based on individual user preferences, fostering more ethical and intelligent user-centric recommendations. To navigate the uncertainty inherent in multi-objective RSs, we develop a Bayesian optimization (BO) framework that captures personalized trade-offs between different objectives while accounting for their uncertain interdependencies. Furthermore, we introduce an orthogonal meta-learning paradigm to enhance BO efficiency and effectiveness by leveraging shared knowledge across similar tasks and mitigating conflicts among objectives through the discovery of orthogonal information. Finally, extensive empirical evaluations demonstrate the effectiveness of our method in optimizing uncertain multi-objectives for individual users, paving the way for more adaptive and user-focused RSs.

\end{abstract}

\settopmatter{printacmref=false}

\maketitle

\section{Introduction}

In today's digital age, recommender systems (RSs)~\cite{zhang2019deep} have become the backbone of information dissemination, revolutionizing the way we access and engage with content. These intelligent systems work tirelessly behind the scenes, analyzing our behaviors and preferences based on historical data to curate personalized information feeds tailored to our tastes and needs. From e-commerce~\cite{liu2024large} and social media~\cite{sun2024self} to education~\cite{zhang2019hierarchical} and healthcare~\cite{cui2022ketch}, RSs, widely investigated in academia and applied in industry~\cite{sun2019research}, have transformed how we discover and consume information, shaping our digital experiences and influencing our decision-making processes.

Early works on RSs mainly focus on improving recommendation accuracy~\cite{sun2023theoretically}. However, the singular focus on accuracy has inadvertently created echo chambers~\cite{yin2023understanding}, where narrowly tailored recommendations confine users to limited information spaces, stifling diversity of thought and experience. As such, more studies have considered comprehensive ethical aspects to enhance the beyond-accuracy performance of RSs~\cite{paparella2023reproducibility}, e.g., diversity~\cite{yin2023understanding}, explanation~\cite{wu2023generic} and fairness~\cite{wu2022multi}. Despite the great success, these methods suffer from a major limitation, i.e., the objectives of optimizing accuracy and beyond-accuracy performance are typically 
combined with pre-defined hyperparameters, indicating all users in RSs share the same objectives. Thus, it fails to reflect real-world complexities, where users may have diverse or uncertain requirements for RSs. For instance, some users may prioritize content diversity, while others might value fairness in their recommendations. 

\begin{figure}[t]
    \centering
    \includegraphics[width=1\linewidth]{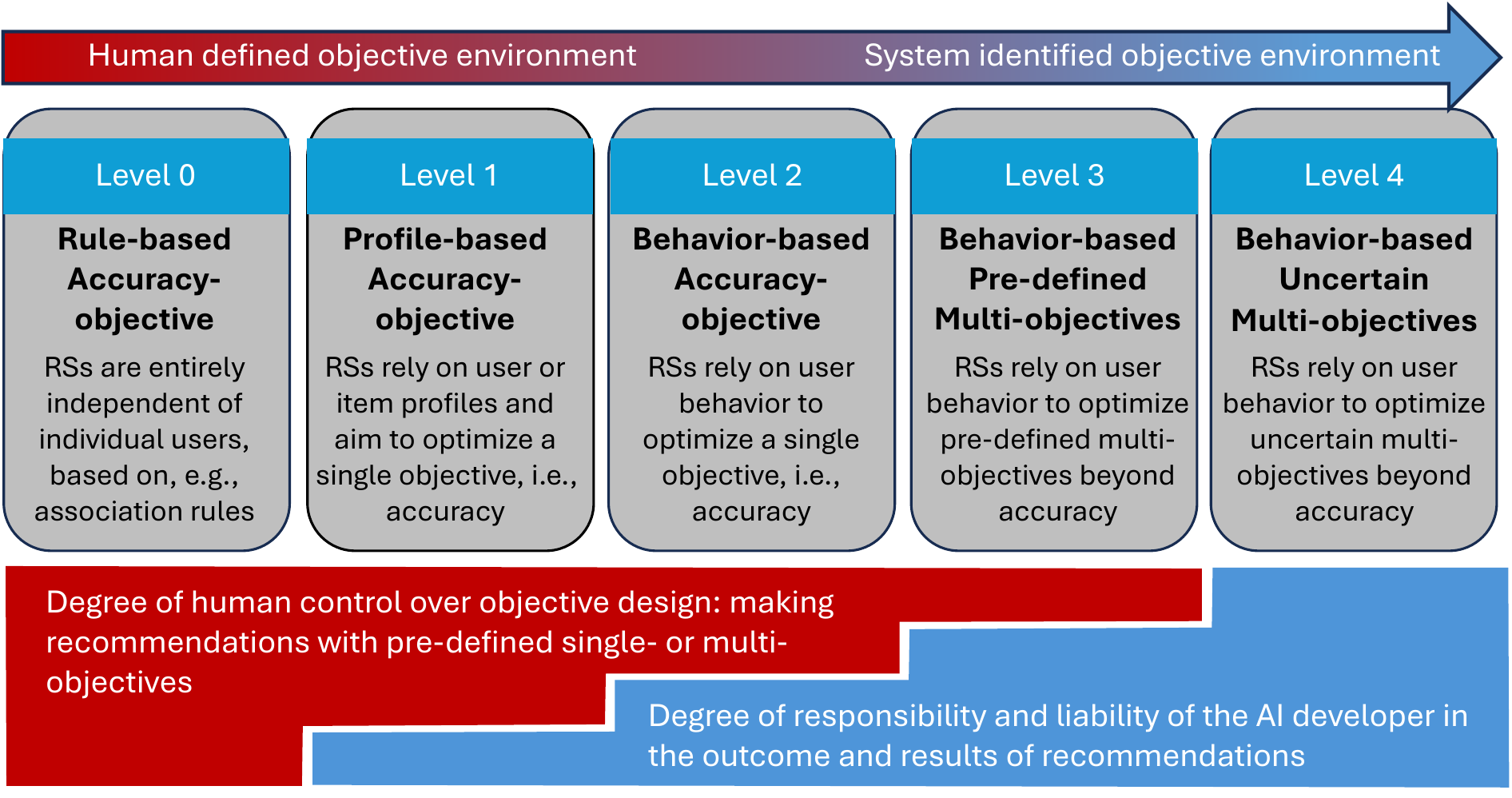}
    \caption{Different levels of autonomy for RSs.}
    \vspace{-0.15in}
    \label{fig:autonomy-level}
    \vspace{-0.15in}
\end{figure}

To elevate user experience and optimize AI's service to humanity, it's imperative to develop more intelligent RSs, which can autonomously adapt to individual user preferences and objectives, offering truly personalized interactions. Drawing parallels with autonomous driving~\cite{dai2024vistarag}, we first propose a novel framework that defines distinct levels of autonomy for RSs based on their ability to independently determine and pursue recommendation objectives. Overall, there are five different levels.

\begin{itemize}[leftmargin=*]
    \item \textit{Level 0: Rule-based Accuracy-objective}. RSs are entirely independent of individual user preference but built upon pre-defined or extracted rules according to the statistical interaction data, such as item popularity or association rules. 
    \item \textit{Level 1: Profile-based Accuracy-objective}. RSs rely on static user or item profiles to generate recommendations (aka. content-based RSs~\cite{sun2019research}) by optimizing a single accuracy-oriented objective.
    \item \textit{Level 2: Behavior-based Accuracy-objective}. RSs use historical personalized user behaviors to make recommendations (aka. collaborative filtering~\cite{sun2019research}), optimizing the accuracy-oriented objective. 
    \item \textit{Level 3: Behavior-based Pre-defined Multi-objectives}. RSs use historical personalized user behaviors to make recommendations that optimize pre-defined multiple (i.e., accuracy and beyond-accuracy) objectives, without considering personalized user needs.  
    \item \textit{Level 4: Behavior-based Uncertain Multi-objectives}. RSs use historical personalized user behaviors to make recommendations that optimize uncertain multiple objectives, i.e., the importance of different objectives is automatically learned by considering personalized user needs, instead of pre-defined hyperparameters.  
\end{itemize}

In this paper, our goal is to build a more intelligent RS at Level 4, automatically modeling the importance of different objectives by considering personalized user needs to improve the overall performance of multiple objectives. Intuitively, assigning personalized weights of objectives to users is a straightforward solution to improve the overall performance of multi-objectives. 
For example, we should lower the weight of the diversity objective in multi-objective learning if a user shows a narrow interest, because blindly increasing diversity may largely harm other objectives such as recommendation accuracy. However, there remain challenges in determining the appropriate weights in multi-objective recommendation quantitatively. 

First, assigning empirical weights (e.g., measured by users' historical behaviors) can not guarantee the desired multi-objective trade-offs in RSs. For ease of illustration, let $l_{uo}(\Theta)$ and $P_{uo}(\Theta)$ denote the recommendation loss (e.g., BPR loss~\cite{sun2022daisyrec}) and the performance (e.g., NDCG~\cite{yin2023understanding}) of a specific objective $o$ (e.g., accuracy) for the user $u$, respectively.  
Specifically, even if we have $\min l_{uo}(\Theta) \leftrightarrow \max P_{uo}(\Theta)$ for each of the $O$ different objectives, 
optimization their combination with empirical weights may not guarantee the optimal performance of multi-objectives, given by, 
\begin{equation}\label{eq1}
    \min \sum\nolimits_{o=1}^{O} \lambda^{emp}_{uo} \cdot l_{uo}(\Theta) \nleftrightarrow \max\sum\nolimits_{o=1}^{O}\lambda^{emp}_{uo} \cdot P_{uo}(\Theta),
\end{equation}
where $\lambda^{emp}_{uo}$ is the empirical weight.
It mainly lies in multi-objectives 
may conflict with each other and their optimization is essentially achieved by proxy losses, leading to the uncertain relationship between the assigned weights and the performance of multi-objectives. 
Secondly, learning trainable weights (e.g., learn weights through overall loss) 
may lead to the degradation of certain objectives, i.e., 
\begin{equation}\label{eq2}
    \min \sum\nolimits_{o=1}^{O} \lambda_{uo}(\Theta) \cdot l_{uo}(\Theta) \nleftrightarrow \max\sum\nolimits_{o=1}^{O} P_{uo}(\Theta),
\end{equation}
where $\lambda_{uo}(\Theta)$ is the trainable weight. This may lead to trivial solutions for multi-objective learning, that is, a lower loss $l_{uo}(\Theta)$ gets a larger weight $\lambda_{uo}(\Theta)$. Thus, some objectives may dominate others, resulting in imbalanced optimization and sub-optimal performance. 

According to Equations (\ref{eq1}) and (\ref{eq2}), the main difficulty lies in the uncertain relationship between the weights and overall objective in multi-objective learning, remaining the black box to determine weights in an empirical or learnable way. To open this black box for autonomous multi-objective learning in RSs, we adapt the Bayesian optimization (BO) to accommodate the personalized needs of individual users, which can efficiently explore the search space in the black box and {quantify uncertainties between the weights and overall objective}.
For each trail of BO, it is typically to train a multi-objective model with specific weights for overall performance measurement. To this end, we propose to accelerate and enhance the training of multi-objective model from two aspects. Firstly, to make use of the correlation between different multi-objective models  for efficient training, 
we propose to utilize meta-learning~\cite{wang2023meta} to facilitate the parameter learning for each new set of aggregation weights, leveraging the shared knowledge across similar optimization tasks. Secondly, to alleviate the conflict among different objectives for effective training, we equip meta-learning with the orthogonal gradient descent strategy to avoid the invalid updating of conflict gradients for better convergence.   

In summary, our main contributions lie three-fold. 
\begin{itemize}[leftmargin=*]
    \item We are the first to propose a novel framework that defines distinct levels of autonomy for RSs based on their ability to independently determine recommendation objectives. Meanwhile, it is also the first trial to open the black box between assigned weights and the overall performance of objectives in multi-objective learning. 
    \item  We propose a novel Bayesian optimization method by boosting \underline{B}ayesian \underline{o}ptimization with an \underline{o}rthogonal \underline{m}eta-\underline{l}earning paradigm, abbreviated as BOOML, to efficiently help optimize the uncertain multi-objective task in RSs. Specifically, it considers the collaborative signals among different multi-objective models for fast convergence and alleviates invalid updating of conflict gradients for better performance.
    \item We conduct empirical studies on three real-world datasetes to demonstrate the effectiveness of our proposed method
    in exploring the uncertain multi-objectives for individual users.   
\end{itemize}

\section{Related Works}
\subsection{RSs at Levels 0-2} 
Early RSs at \textit{Level 0} rely on generic rules or broad statistical patterns, such as recommending the most popular items, or frequently co-occurred items mined by association rules~\cite{sun2019research}, thus failing to provide personalization.
Later, RSs at \textit{Level 1} began leveraging static user or item profiles, aka. content-based RSs~\cite{sun2019research}, for instance, a user who indicates a preference for `romance' in their profile would receive recommendations for romantic movies. Hence, a basic level of personalization is introduced.
Advancing to \textit{Level 1}, RSs at \textit{Level 2} resort to dynamic historical user behaviors to learn user preference, aka. collaborative filtering based RSs~\cite{sun2019research}. Different techniques are adopted, ranging from simple matrix factorization (MF)~\cite{he2016vbpr}, to complex deep learning, e.g., MLP~\cite{sun2024self}, RNN~\cite{sun2023theoretically}, GCN~\cite{peng2024less}, Transformer~\cite{zhang2022next} and LLMs~\cite{liu2024large,wang2025re2llm}.  
However, RSs at Levels 0-2 aim to purely improve recommendation accuracy, ignoring other essential ethical aspects, e.g., diversity and fairness. 
 
\subsection{RSs at Level 3}
RSs at \textit{Level 3} exploit dynamic historical user behaviors to learn user preference by optimizing pre-defined multi-objectives beyond accuracy. As we primarily focus on two key ethical aspects -- diversity and fairness, we limit our discussion to research relevant to these areas. Studies on other ethical aspects, e.g., explanation and privacy-perseveration, will be explored in our future work.  

\smallskip\noindent\textit{\textbf{Diversity}} bias would cause filter bubbles, which grow along the feedback loop and inadvertently narrow user interests~\cite{khenissi2020theoretical}. Thus, a vital branch is to enhance recommendation diversity while maintaining accuracy, mainly divided into three categories: post-processing heuristic methods~\cite{steck2018calibrated,lin2022feature,sha2016framework}, determinantal point process methods~\cite{chen2018fast,wu2019pd,gan2020enhancing,liu2022determinantal,warlop2019tensorized} and end-to-end learning methods~\cite{zheng2021dgcn,liang2021enhancing,chen2020improving,chen2021multi,shi2024diversifying,yang2023dgrec,wang2019modeling,cen2020controllable,lu2021future,wang2024sparks,stamenkovic2022choosing,liu2023generative}. However, they suffer from different limitations: (1) some follow a two-stage paradigm, i.e., train offline models to score items on accuracy and then re-rank items considering diversity; and (2) others incorporate accuracy and diversity objectives with a pre-defined ``trade-off'' hyperparameter, overlooking the uncertainty of personalized user needs. 

\smallskip\noindent\textit{\textbf{Fairness}} is another critical ethical issue of RSs~\cite{wang2023survey,li2023fairness,deldjoo2024fairness} that can affect personal experience and social good since RSs serve a resource allocation role in society by allocating information to users and exposure to items.
Extensive work has encouraged equal exposure across item groups partitioned by item features, such as category and popularity\footnote{
There are different types of fairness in RSs, e.g., user fairness, item fairness, and joint fairness~\cite{wang2023survey}. In this study, we primarily focus on item fairness regarding popularity without relying on extra item features.}. 
Early studies design {data-oriented methods}~\cite{ekstrand2018all} to alleviate the unfairness issue by changing training data. 
Another branch focuses on {re-ranking based methods}~\cite{liu2019personalized,steck2018calibrated} to adjust the outputs of recommendation models to promote fairness. 
Recent studies propose {ranking-based methods} to improve fairness by (1) using linear programming to add fairness constraints~\cite{singh2018fairness}; (2) adding a fairness-related regularization term to the recommendation loss~\cite{beutel2019fairness,zhu2018fairness}; 
(3) leveraging adversarial learning to learn fair representations or predicted scores~\cite{bose2019compositional,wu2021learning,zhu2020measuring}; (4) adopting reinforcement learning to achieve long-term fair recommendations~\cite{ge2021towards}; and 
(5) balancing accuracy and fairness for various stakeholders with heuristic strategies~\cite{wu2021tfrom,patro2020fairrec} or Pareto optimality guarantee~\cite{wu2022multi,ge2022toward}. 
Despite the effectiveness, most of them mainly seek a uniform ``trade-off" between accuracy and fairness across all users while ignoring personalized user needs.  

\subsection{RSs at Level 4}
Some studies attempt to achieve RSs at \textit{Level 4}. For instance, in \cite{wang2022user}, the authors propose a new recommender prototype called User Controllable RS, which enables users to
actively control the mitigation of filter bubbles. Nevertheless, it relies on user feedback and only considers the balance between accuracy and diversity. 
MMoE~\cite{ma2018modeling} adapts the Mixture-of-Experts structure to multi-task learning by sharing the expert submodels across all tasks, while also having a gating network to optimize each task. However, it only learns the gates at the task level instead of the individual user level. A recent work on arXiv~\cite{li2024deep} introduces a deep Pareto reinforcement learning model for multi-objective RSs, which accounts for the relationships between different objectives and implements personalized dynamic weighting for these objectives. However, it still relies on learning trainable weights for multiple objectives, leading to the degradation of certain objectives. Besides, it ignores the potential conflicts of different objectives and introduces substantial computational complexity due to dynamically adjusting objective weights based on individual user information. 

\begin{table*}[t]
\footnotesize
\centering
\setlength\tabcolsep{5.5pt}
\caption{Performance of empirical weights on Amazon-Games. Similar trends are noted on all datasets in our study.}\label{tab:empirical-weights}
\vspace{-0.15in}
\begin{tabular}{|c|ccc|ccc|ccc|ccc|ccc|}
    \specialrule{.15em}{.05em}{.05em}
    \multirow{2}{*}{$K=50$}
    &\multicolumn{3}{c|}{$\beta = 0.1$} &\multicolumn{3}{c|}{$\beta = 0.5$}
    &\multicolumn{3}{c|}{$\beta = 1.0$} &\multicolumn{3}{c|}{$\beta = 5.0$}&\multicolumn{3}{c|}{$\beta = 10$}\\ \cline{2-16}
    &NDCG &ILD &ARP &NDCG &ILD &ARP &NDCG &ILD &ARP &NDCG &ILD &ARP &NDCG &ILD &ARP \\
    \specialrule{.05em}{.05em}{.05em}
    $\lambda= 0.1$  &\cellcolor{red!25}0.0751&0.9800&68.5387 &0.0744&1.0163&72.0956 &0.0695&0.9599&69.9408 &0.0710&1.0455&69.9361 &0.0656&\cellcolor{blue!25}1.0518&68.5614 \\
    $\lambda= 0.5$  &0.0668&0.9520&67.9519 &0.0691&0.8386&70.8955 &0.0721&0.8182&70.1282 &0.0708&0.9428&69.8305 &0.0717&0.8452&75.4949 \\
    $\lambda= 1.0$  &0.0680&0.8756&71.2285 &0.0655&0.8659&70.4169 &0.0627&0.8099&67.0925 &0.0704&0.8172&72.5754 &0.0674&0.8426&69.9872 \\
    $\lambda= 5.0$  &0.0621&0.5722&63.3405 &0.0584&0.5466&63.1538 &0.0594&0.5478&65.0404 &0.0648&0.6063&66.4698 &0.0638&0.6197&68.7281 \\
    $\lambda= 10$  &0.0594&0.6007&65.1894 &0.0656&0.5195&65.4786 &0.0610&0.5871&64.1471 &0.0566&0.5663&\cellcolor{green!25}63.1400 &0.0613&0.6395&64.9703 \\
    \specialrule{.15em}{.05em}{.05em}
    \end{tabular}
    \vspace{-0.1in}
\end{table*}

\section{Uncertain Multi-Objectives}
This section first introduces different objectives in RSs by considering accuracy and different ethics, followed by the formulation of our uncertain multi-objective function.
In this paper, we focus on three objectives without loss of generality, including accuracy, diversity, and fairness. Note that, our framework can be easily adopted and adapted to more objectives.

\smallskip\noindent\textbf{Notations}. Let $\mathcal{U} = \{u_1, u_2, \ldots, u_{|\mathcal{U}|}\}$, $\mathcal{V} = \{v_1, v_2, \ldots, v_{|\mathcal{V}|}\}$ and $\mathcal{C} = \{c_1, c_2, \ldots, c_{|\mathcal{C}|}\}$ denote the user, item and item category sets, respectively. 
$\bm{R} \in \mathbb{R}^{|\mathcal{U}| \times |\mathcal{V}|}$ denotes the user-item interaction matrix, where its entries $r_{ij}=1$ represents user $u_i$ interacted with item $v_j$; otherwise 0.
For each item $v_j$, it has a categorical feature $c(v_j) \in \mathcal{C}$. 
To model users and items in the latent space, we embedding them into the user representation matrix  $\bm{U} \in \mathbb{R}^{|\mathcal{U}| \times d}$ and the item representation matrix $\bm{V} \in \mathbb{R}^{|\mathcal{V}| \times d}$, where $d$ is dimension of the latent space. 

\smallskip\noindent\textbf{Problem Statement}. Given the user-item interaction $\mathbf{R}$, our goal is to provide a personalized recommendation list (RL) with the ranking of $K$ items to each user, aiming to better hit her preference while meeting her personalized requirements regarding different ethical aspects, e.g., diversity and fairness. 

\subsection{Different Objectives and Metrics}
\label{sec:objectives}
\smallskip\noindent\textbf{Accuracy Objective}. The primary goal of RSs is to provide accurate recommendations to hit user preference (e.g., ground-truth interacted items). The accuracy can be measured with widely-used ranking metrics, e.g., Precision, Recall, and NDCG~\cite{sun2022daisyrec}. In our study, we adopt NDCG as the evaluation metric, denoted as ACC, as it evaluates whether (1) the target items are correctly recommended and (2) the correctly recommended items are top-ranked. Larger values of NDCG indicate better ranking accuracy. 

\smallskip\noindent\textbf{Accuracy Optimization}. We adopt the BPR loss~\cite{sun2022daisyrec} to maximize the preference gap between positive and negative items for all users, 
\begin{equation}
\small
f_{\text{acc}}(\bm{\Theta}) = - \sum\nolimits_{(u_i,v_j,v_k) \in \mathcal{D}_T} \log \sigma(\hat{r}_{ij} - \hat{r}_{ik}),
\end{equation}
where $\hat{r}_{ij} = \bm{u}_i^T\bm{v}_j$ is the estimated preference score of user $u_i$ to item $v_j$; $\bm{u}_i$ and $\bm{v}_j$ denote the encoding of user $u_i$ and item $v_j$, respectively; $\mathcal{D}_T$ denotes the training set meaning $u_i$ engaged $v_j$ instead of $v_k$, i.e., $r_{ij}=1$ and $r_{ik}=0$; and $\sigma(x) = {1}/(1 + \exp(-x))$ is the sigmoid function.

\smallskip\noindent\textbf{Diversity Objective}. 
To alleviate filter bubbles \cite{wang2024sparks}, it is necessary to provide diversified recommendations rather than focusing narrowly on specific categories of items. Typically, the recommendation diversity can be measured with pairwise diversity metrics, e.g., ILD (intra-list distance), entropy-and-diversity score~\cite{yin2024simple}. In our method, we adopt ILD to measure the average Euclidean distance between every pair of items
in the RL, i.e.,
%
\begin{equation}\small
      DIV =\frac{1}{|\mathcal{U}|}\sum\nolimits_{u_i\in\mathcal{U}}\sum\nolimits_{(v_j, v_k)\in RL_{u_i}, v_j\neq v_k}\frac{||\bm{v}_j - \bm{v}_k||_2}{\vert RL_{u_i}\vert\times(\vert RL_{u_i}\vert-1)},
\end{equation}
where $RL_{u_i}$ denotes the recommendation list (RL) for user $u_i$.
A larger value of IDL indicates a more diverse result in the RL.

\smallskip\noindent\textbf{Diversity Optimization}. In our study, we propose to maximize the diversity measured by the negative entropy of estimated category probability distribution for all users as in~\cite{yin2024simple},
\begin{equation}
\small
f_{{div}}(\bm{\Theta}) = -\sum\nolimits_{u_i\in \mathcal{U}} Entropy(\hat{p}_i) = \sum\nolimits_{u_i\in \mathcal{U}}\sum\nolimits_{l=1}^{|\mathcal{C}|} \hat{p}_{il} \log \hat{p}_{il},
\end{equation}
where $\hat{p}_i$ is the estimated category probability distribution for user $u_i$, satisfying $\sum_l \hat{p}_{il} = 1$; and $\hat{p}_{il}$ denotes user $u_i$'s preference towards category $c_l$. 
Specifically, $\hat{p}_{il}$ can be estimated by aggregating $u_i$'s preference towards all items belonging to category $c_l$, 
\begin{equation}
\small
     \hat{p}_{il} = Softmax\left(\sum\nolimits_{v_j\in \mathcal{V}} \mathbb{I}(c(v_j) =c_l)\cdot \hat{r}_{ij}\right),
\end{equation}
\noindent where $ \mathbb{I}(\cdot)$ denotes the indicator function. The Softmax function making it a probability distribution, ensuring non-negativity $\hat{p}_{il}\ge0$ and $\sum_l \hat{p}_{il} = 1$ for $\hat{p}_i$.

\begin{figure}[t]
    \centering
    \subfigure[Total Loss]{
        \includegraphics[width=0.31\linewidth]{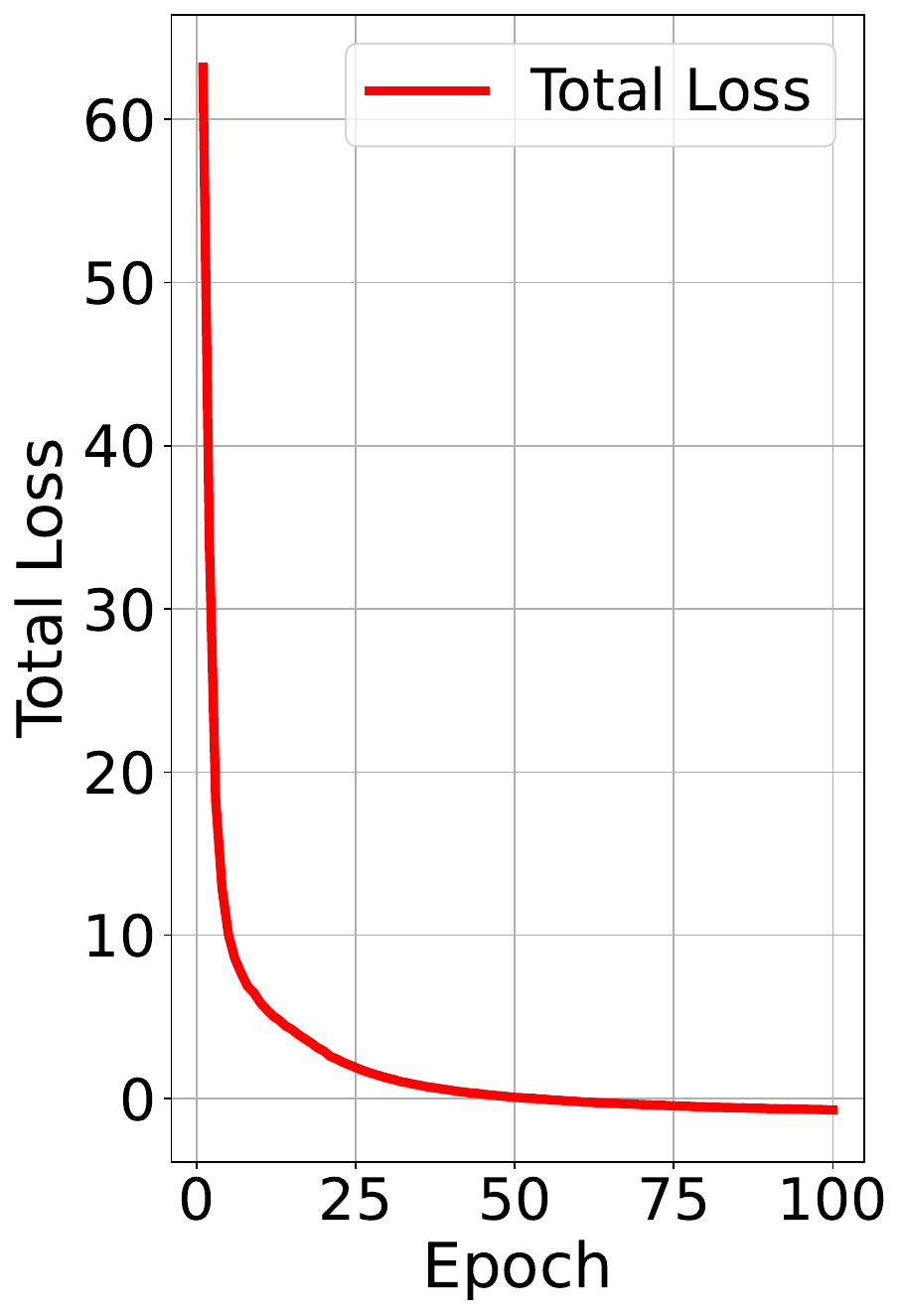}
    }
    \hspace{-0.05in}
    \subfigure[Weights]{
        \includegraphics[width=0.31\linewidth]{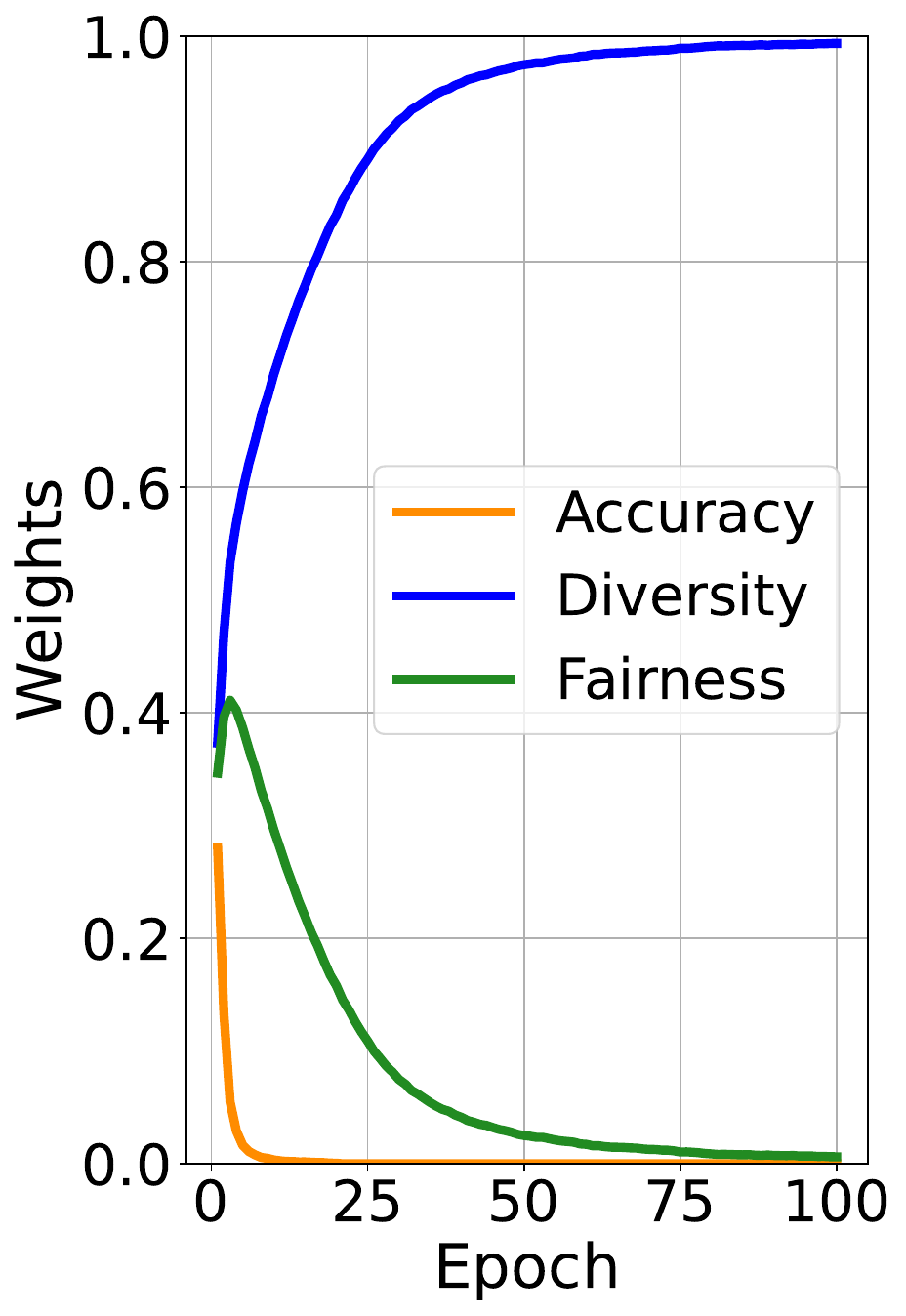}
    }
    \hspace{-0.05in}
    \subfigure[Separate Loss]{
        \includegraphics[width=0.31\linewidth]{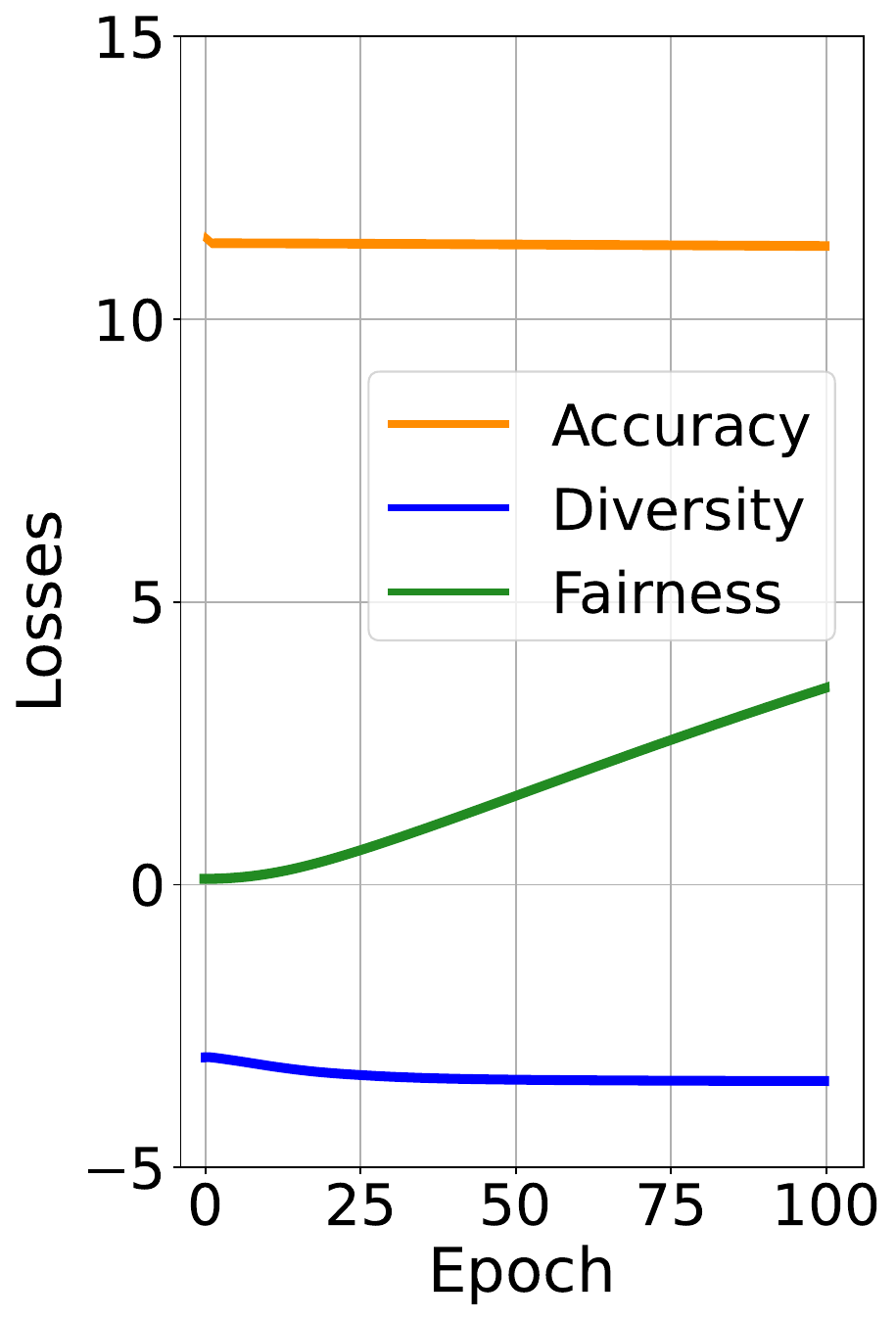}
    }
    \vspace{-0.2in}
    \caption{Performance of trainable weights on Games. Similar trends are noted on other datasets in our study.}
    \label{fig:sgd-performance}
    \vspace{-0.1in}
\end{figure}

\smallskip\noindent\textbf{Fairness Objective}. Fairness aims to ensure the recommendation results are not dominated by popular products but include long-tail items~\cite{gupta2021causer}. The recommendation fairness regarding popularity can be measured by several metrics, e.g., ARP (Average Recommendation Popularity)~\cite{gupta2021causer}, RR (Recommendation Rate)~\cite{zhang2021causal}, and PR (Popularity Rate)~\cite{ge2021towards}.  For generality, we use ARP to measure the average popularity of the recommended items, i.e.,
\begin{equation}
\small
    FAIR = \frac{1}{|\mathcal{U}|}\sum\nolimits_{u_i\in\mathcal{U}}\sum\nolimits_{v_j \in RL_{u_i}} \frac{\phi(v_j)}{\vert RL_{u_i}\vert},
\end{equation}
\noindent where $\phi(v_j)$ represents the popularity of item $v_j$. 
Smaller values of ARP indicate fairer recommendation results.

\smallskip\noindent\textbf{Fairness Optimization}. Intuitively, since popular items are more frequently interacted with by users, their representations are likely to be pulled closer to user representations during the model training process, leading to systematic higher scores. Inspired by Biased-MF~\cite{koren2009matrix}, we propose to remove such bias by minimizing the gap between the estimated preference score of individual users over individual items and the estimated average score of the system, 
\begin{equation}
\small
    f_{{fair}}(\bm{\Theta}) = \frac{1}{|\mathcal{U}||\mathcal{V}|} \sum\nolimits_{i=1}^{|\mathcal{U}|}\sum\nolimits_{j=1}^{|\mathcal{V}|} \left|\sigma(\hat{r}_{ij}) -\overline{r}\right|,
\end{equation}
where $\overline{r} = \sum\nolimits_{i=1}^{|\mathcal{U}|}\sum\nolimits_{j=1}^{|\mathcal{V}|} \sigma(\hat{r}_{ij})/{(|\mathcal{U}|\cdot|\mathcal{V}|})$ is the average predicted score for all users towards all items.

\subsection{Uncertain Multi-Objectives}\label{subsec:uncertain-objectives}
In this paper, we aim to improve the overall performance of multi-objectives, while keeping validation of each objective, i.e.,
\begin{equation}\label{goal}
\small
\begin{aligned}
    &\max_{\bm{\Theta}} g(ACC,DIV,FAIR), \\
    \textit{s.t.} \quad ACC&>\tau_{acc}, DIV>\tau_{div},FAIR<\tau_{fair},
\end{aligned}   
\end{equation}
where $g(\cdot)$ denotes the overall performance of multiple objectives; and $\tau_{acc}$, $\tau_{div}$, and $\tau_{fair}$ represent the thresholds of minimal requirement for accuracy, diversity, and fairness objectives, respectively.  

In real-world scenarios, users may have diverse or uncertain requirements in RSs, leading to varying importance in optimizing multiple objectives for different users. For example, if a user shows a narrow interest in items, blindly increasing recommendation diversity may largely harm other objectives such as recommendation accuracy. To this end, we propose to optimize the personalized multi-objectives to capture users' uncertain requirements in RSs, enabling RSs to function as more ethical and intelligent user-centric assistants. Specifically, we assign personalized weights for different objective losses for multi-objective optimization to improve the overall performance of multi-objectives,
\begin{equation}\label{equ:uncertain-multi-objective}
\small
    \mathcal{F}(\bm{\lambda}, \bm{\beta}) = \sum\nolimits_{u_i \in \mathcal{U}}[f_{acc}(\bm{\Theta}_i) + \lambda_i f_{div}(\bm{\Theta}_i) + \beta_i f_{fair}(\bm{\Theta}_i)], 
\end{equation}
where $\lambda_i$ and $\beta_i$ are the personalized weights of diversity and fairness objectives for $u_i$.
However, challenges persist in quantitatively determining the appropriate weights using existing methods.

\smallskip\noindent\textbf{Why Not Empirical Weights}?
Assigning empirical weights via the grid search for different objectives~\cite{chen2020improving,wang2024sparks,beutel2019fairness,zhu2018fairness}
has been widely used in multi-objective learning due to its simplicity 
for RSs at Level 3. However, for RSs at Level 4, the scale of grid search is exponential to the size of objectives and users, leading to unacceptable costs in the training phase. Worsely, it is intractable to clarify the certain relationship between weights and multi-objective performance through empirical investigation. Table~\ref{tab:empirical-weights} illustrates the impact of weight changes on multi-objective performance (MF as encoder), where we apply a grid search in $\{0.1, 0.5, 1.0, 5.0, 10\}$ for the weights of diversity ($\lambda$) and fairness ($\beta$). The optimal performance for different objectives is highlighted in different colors.   
We observe that the optimal performance for diversity (in blue) is achieved with a smaller weight on diversity ($\lambda=0.1$) but a larger weight on fairness ($\beta=10$).
This indicates \textit{the uncertain relationship between weights and performance of multi-objectives makes it hard to find the optimal weights for multi-objective learning.}

\smallskip\noindent\textbf{Why Not Trainable Weights}?
Some methods attempt to learn trainable weights that aggregate multiple objectives for unified learning~\cite{li2024deep}, however, it may lead to the degradation of certain objectives. For example, a trivial solution that assigns a lower loss 
with a larger weight,
results in imbalanced optimization and sub-optimal performance. Figure~\ref{fig:sgd-performance} depicts the multi-objective performance of learning trainable weights with SGD using MF as encoder across different training epochs on Amazon-Games. We note that the total loss decreases rapidly until reaching a small value as in Figure~\ref{fig:sgd-performance}(a), indicating the optimization process converges. However, the weights for accuracy and fairness drop significantly as the epochs increase shown in Figure~\ref{fig:sgd-performance}(b), leading to diversity dominating the optimization process, i.e., only the loss for diversity decreases to a small value, whereas the losses for accuracy and fairness remain relatively high as shown in Figure~\ref{fig:sgd-performance}(c). This validates that \textit{learning trainable weights with SGD cannot adequately balance different objectives to achieve optimal recommendation performance}.

\section{Bayesian Optimization Boosted via Orthogonal Meta-Learning}
Guided by the analysis in Section~\ref{subsec:uncertain-objectives}, it is hard to determine optimal weights for multi-objective learning through empirical investigation or direct optimization. To explore the uncertain relationships between the weights and multiple objectives, we propose a novel Bayesian optimization method to open the black box that achieves balanced optimization among different objectives and bridges the gap between objective losses and performances. Most importantly, for more efficient and effective optimization, we design an orthogonal meta-learning paradigm to enhance the optimization of each objective by considering their correlations and potential conflicts.

\subsection{Bayesian Optimization for Group-Level Personalization}

\subsubsection{Group-Level Uncertain Multi-Objectives}

Recall Equation~(\ref{equ:uncertain-multi-objective}), it is impractical to directly leverage Bayesian optimization to find out the optimal $\lambda_i$ and $\beta_i$ for each user $u_i$, as the search space is huge due to the large volume of users in RSs. To this end, we allocate users into different groups based on the statistics of their behaviors, as similar users may share a similar need (e.g., tendency toward diversity and fairness) for items. Specifically, we utilize three kinds of user behavior statistics, including the total number of engaged items, the ratio of engaged categories to items, and the average popularity of engaged items, which could reflect users' preferences towards diversity and fairness of recommendation results. Thus, we cluster users into $W$ different groups ($\{\mathcal{G}_1,\cdots,\mathcal{G}_W\}$) based on these statistical features. For each group $\mathcal{G}_w$,  we assign a personalized parameter pair $(\lambda_w,\beta_w)$ for multi-objective learning. Accordingly, the group-level uncertain multi-objective function is given by: 
%
\begin{equation}
\small
\begin{aligned}
    &\mathcal{F}^{\bm{\lambda}, \bm{\beta}}(\bm{\Theta}) =  \sum\nolimits_{i=1}^{|\mathcal{U}|}\mathcal{F}_i^{\bm{\lambda}, \bm{\beta}}(\bm{\Theta_i}), \\
   & \mathcal{F}_i^{\bm{\lambda}, \bm{\beta}}(\bm{\Theta_i}) = \sum\nolimits_{w=1}^{W}\mathbb{I}(u_i\in \mathcal{G}_w)\cdot[f_{acc}(\bm{\Theta}_i) +  \lambda_w f_{div}(\bm{\Theta}_i) + \beta_w f_{fair}(\bm{\Theta}_i)],
\end{aligned}
\end{equation}
where $\mathbb{I}(u_i\in \mathcal{G}_w)$ aims to select $\lambda_w$ and $\beta_w$ for user $u_i$; $\bm{\lambda} = [\lambda_1,\cdots,\lambda_W]$ and $\bm{\beta} = [\beta_1,\cdots,\beta_W]$; and {$\bm{\Theta} = [\Theta_1,\cdots,\Theta_{|\mathcal{U}|}]$}, where $\Theta_i$ denotes the learnable parameters related to user $u_i$ and her engaged items.
Hence, our goal is to find out the optimal $\lambda_w$ and $\beta_w$ for each group $\mathcal{G}_w$, thus satisfying users' uncertain requirements regarding various ethical aspects at the group-level.

\subsubsection{Bayesian Optimization}\label{sec:BO}

For optimal weights $\bm{\lambda}$ and $\bm{\beta}$, we formulate Equation (\ref{goal}) as a Bayesian optimization (BO) problem,
\begin{equation}
\small
\begin{aligned}
    \max_{\bm{\lambda,\beta}} g(ACC(\Theta^{\bm{\lambda,\beta}}),&DIV(\Theta^{\bm{\lambda,\beta}}),FAIR(\Theta^{\bm{\lambda,\beta}})) - \kappa\cdot const(\Theta^{\bm{\lambda,\beta}}),\\
    & \Theta^{\bm{\lambda,\beta}} = \min_\Theta \mathcal{F}^{\bm{\lambda}, \bm{\beta}}(\Theta),
\end{aligned}
\end{equation}
where $\Theta^{\bm{\lambda,\beta}}$ denotes the solution of multi-objective function $\mathcal{F}^{\bm{\lambda}, \bm{\beta}}(\Theta)$ with weights $\bm{\lambda}$ and $\bm{\beta}$. The soft constraint $const(\Theta^{\bm{\lambda,\beta}}) = \lfloor\tau_{acc}-ACC(\Theta^{\bm{\lambda,\beta}})\rfloor_+ + \lfloor\tau_{div}-DIV(\Theta^{\bm{\lambda,\beta}})\rfloor_+ + \lfloor FAIR(\Theta^{\bm{\lambda,\beta}}) - \tau_{fair}\rfloor_+$ penalize the unsatisfied constraints in Equation (\ref{goal}) with a penalty coefficient $\kappa>>0$. For the function $g(\cdot)$, we define the overall performance of multiple objectives in two ways:

\begin{itemize}[leftmargin=*]
    \item \textit{Rescaled Sum}. It seeks the maximal sum of different objectives. 
    However, measuring objectives with different metrics usually has different scales, e.g., $NDCG$ $\in [0,1]$, whereas $ILD$ may be larger than 1 and $ARP$ possess the opposite trend with $NDCG$ and $ILD$ (i.e., smaller ARP values indicate fairer recommendation). To this end, we adopt the rescaled sum to formulate 
    $g(NDCG,ILD,ARP) = NDCG + \sigma(ILD) + \sigma(1/ARP)$.   
    \item \textit{Harmonic Mean}. It seeks the maximal harmonic mean of different objectives. Considering the opposite trend of $ARP$ compared with $NDCG$ and $ILD$, we, therefore, formulate the harmonic mean of these three metrics as $g(NDCG,ILD,ARP)  = 3/[NDCG^{-1} + \sigma(ILD)^{-1} + \sigma(1/ARP)^{-1}]$. 
\end{itemize}

Following the standard procedure of BO, we iteratively update a surrogate model to approximate the objective function \( g(\cdot) \) and guide the search for optimal weights 
$\bm{\lambda}$ and $\bm{\beta}$.
Specifically, the procedure includes the following steps:
We start by selecting an initial set of points (where a point is a combination of $\bm{\lambda}, \bm{\beta}$) and evaluate the objective function. A Gaussian process surrogate model is then fitted to approximate the objective. We adopt expected improvement as an acquisition function $EI(\cdot)$ to balance exploration (searching unexplored regions) and exploitation (refining known promising areas), where points with high expected improvement are more likely to be sampled
as the next candidate point added to the training data. This process iterates, refining the surrogate model and optimizing the acquisition function, until a convergence criterion is met or the search budget is exhausted.

\subsection{Orthogonal Meta-Learning for Efficient and Effective Optimization}

Each acquisition in BO requires a whole process of multi-objective learning, leading to high cost if each acquisition is conducted independently. To this end, we propose an efficient and effective training optimization for two aspects, namely meta optimization and orthogonal gradient descent. The meta optimization can reduce the times of gradient updating by exploiting shared knowledge across similar tasks, leading to efficient optimization to a new task. The orthogonal gradient descent can alleviate the conflict among different objectives, therefore further improving the effectiveness of the meta optimization.

\begin{algorithm}[t]
\small
\caption{BOOML}\label{alg:booml}
\LinesNumbered
\KwIn{
    Support set $\mathcal{S}$, query set $\mathcal{Q}$, max trails of Bayesian optimization $T$ and meta-learning epoch $E_{ml}$\\
    
}
\KwOut{Weight vectors $\boldsymbol{\lambda}, \boldsymbol{\beta}$, and model parameters $\boldsymbol{\Theta}$}
Initialize a random set $\{(\boldsymbol{\lambda}, \boldsymbol{\beta})_0, (\boldsymbol{\lambda}, \boldsymbol{\beta})_1, \dots, (\boldsymbol{\lambda}, \boldsymbol{\beta})_k\}$ \\
$\mathcal{I}$ = [ ]\;
\For{$t = 1$ to $T$}{
    \If{$t \leq k$}{
            Select $(\bm{\lambda}, \bm{\beta})_t$ from the initialized set\;
        }
    \Else{
            $(\boldsymbol{\lambda}, \boldsymbol{\beta}) = \arg\max_{(\boldsymbol{\lambda}, \boldsymbol{\beta})} \mathrm{EI}(\mathcal{I})$\tcp*{Expected improvement}
    }
        \For{$e = 1$ to $E_{ml}$ }{
            \For{$u_i$ involved in $\mathcal{S}$ }{
                $\boldsymbol{\Theta}_i' = \boldsymbol{\Theta}_i - \eta_1 \nabla_{\boldsymbol{\Theta}_i} \mathcal{F}_i^{\boldsymbol{\lambda}, \boldsymbol{\beta}}(\mathcal{S}_i, \boldsymbol{\Theta}_i)$\tcp*{Inner loop}
            }
            
            \For{$u_i$ involved in $\mathcal{Q}$ }{
                $\bm{g}_{o_m} = \nabla_{\boldsymbol{\Theta}_i} f_{o_m}(\mathcal{Q}_i, \boldsymbol{\Theta}_i - \eta_1 \nabla_{\boldsymbol{\Theta}_i} \mathcal{F}^{\boldsymbol{\lambda}, \boldsymbol{\beta}}(\mathcal{S}_i, \boldsymbol{\Theta}_i))$\;  
                $\tilde{\bm{g}}_{o_m} = \bm{g}_{o_m} - \sum_{o_n\neq {o_m}}\frac{min(\langle \bm{g}_{o_m}, \bm{g}_{o_n} \rangle,0)}{\|\bm{g}_{o_n}\|^2} \bm{g}_{o_n}$\;
                $\boldsymbol{\Theta}_i \gets \boldsymbol{\Theta}_i - \eta_2 \cdot\sum_{o_m}\tilde{\bm{g}}_{o_m}$ \tcp*{Outer loop}
            }

        $\xi= g(ACC(\Theta^{\bm{\lambda,\beta}}),DIV(\Theta^{\bm{\lambda,\beta}}),FAIR(\Theta^{\bm{\lambda,\beta}})) $\;
        $\mathcal{I}.append(\xi,\bm{\lambda,\beta})$\;
        }
    }
$(\boldsymbol{\lambda}^*, \boldsymbol{\beta}^*, \boldsymbol{\Theta}^*) = \arg\max_{(\boldsymbol{\lambda}, \boldsymbol{\beta}) \in \mathcal{I}} \xi$\;
\Return{$\boldsymbol{\lambda}^*, \boldsymbol{\beta}^* ,\boldsymbol{\Theta}^*$}
\end{algorithm}

\subsubsection{Meta Optimization}
To optimize the model effectively, we integrate group correlation and collaborative information into the meta optimization process, enabling the model to generalize better across users by leveraging shared patterns. Specifically, the meta optimization process involves two critical steps: inner loop optimization and outer loop validation, designed to achieve fast adaptation and balance between objectives. To optimize model parameters and validate performance, we divide behaviors of user $u_i$ into a support set $\mathcal{S}_i$ and a query set $\mathcal{Q}_i$.
 
\textbf{For inner loop optimization (support set training)}, we optimize the parameters $\bm{\Theta}_i$ on the support set ($\mathcal{S}_i$) by minimizing the group-level multi-objective loss for each user $u_i$. The updated parameters are computed as:
\begin{equation}
\small
\begin{aligned}
    &\bm{\Theta}_i' = \bm{\Theta}_i - \eta_1 \nabla_{\bm{\Theta}_i} \mathcal{F}_i^{\bm{\lambda}, \bm{\beta}}(\mathcal{S}_i, \bm{\Theta}_i), \\
     \mathcal{F}_i^{\bm{\lambda}, \bm{\beta}}(\mathcal{S}_i, \bm{\Theta}_i) = &f_{acc}(\mathcal{S}_i,\bm{\Theta}_i) +  \lambda_w f_{div}(\mathcal{S}_i,\bm{\Theta}_i) + \beta_w f_{fair}(\mathcal{S}_i,\bm{\Theta}_i),
\end{aligned}
\end{equation}
where $\eta_1$ is the learning rate, and $\mathcal{F}_i^{\bm{\lambda}, \bm{\beta}}(\mathcal{S}_i, \bm{\Theta}_i)$ represents the multi-objective loss function for the support set of user $u_i$. This step leverages group-level personalized weights $(\lambda_w, \beta_w)$ to capture user-specific multi-objective preferences.

\textbf{For outer loop validation (query set evaluation)}, we evaluate the model's generalization based on the meta-loss on the query set ($\mathcal{Q}_i$) using the updated parameters $\bm{\Theta}_i'$:
\begin{equation}
\small
    \mathcal{L}_{\text{meta}, i} = \mathcal{F}_i^{\bm{\lambda}, \bm{\beta}}(\mathcal{Q}_i, \bm{\Theta}_i - \eta_1 \nabla_{\bm{\Theta}_i} \mathcal{F}_i^{\bm{\lambda}, \bm{\beta}}(\mathcal{S}_i, \bm{\Theta}_i)).
\end{equation}
By comparing the performance across several users, shared patterns can be identified for ensuring that the model effectively leverages group and collaborative information.

To balance optimization across all users, we aggregate their meta-loss, i.e., $\mathcal{L}_{\text{meta}} = \sum\nolimits_{i=1}^{\vert\mathcal{U}\vert} \mathcal{L}_{\text{meta}, i}$.
Finally, the global parameters $\bm{\Theta}$ are updated to improve the model’s performance across all tasks,
\begin{equation}
\small
    \bm{\Theta}_i = \bm{\Theta}_i - \eta_2 \nabla_{\bm{\Theta}_i} \mathcal{L}_{\text{meta}}.
\end{equation}

\subsubsection{Orthogonal Gradient Descent}
As users have different objectives within each group, we aim to alleviate conflicts among objectives and improve the effectiveness of meta-learning. 
For instance, increasing diversity may exacerbate fairness, that is, RSs may recommend more popular items in each category~\cite{wang2023survey}. 
To address this issue, we introduce an orthogonal gradient approach to alleviate the conflict among different objectives for outer loop gradient updating, involving gradient computation and adjustment by PCGrad~\cite{yu2020gradient}. First, we calculate the gradient for each objective, 
%
\begin{equation}
\small
    \bm{g}_{o_m}= \nabla f_{o_m}(\mathcal{Q}_i,\bm{\Theta}_i - \eta_1 \nabla_{\bm{\Theta}_i} \mathcal{F}_i^{\bm{\lambda}, \bm{\beta}}(\mathcal{S}_i, \bm{\Theta}_i)),
\end{equation}
where $o_m\in\{acc,div,fair\}$. Then, we detect conflicts between pairs of task gradients  $\bm{g}_{o_m}$ and $\bm{g}_{o_n}$ (e.g., $\bm{g}_{acc}$ and $\bm{g}_{div}$) by comparing their inner product, i.e., $\langle \bm{g}_{o_m}, \bm{g}_{o_n} \rangle < 0 \Rightarrow \text{conflict between } \bm{g}_{o_m} $ and $\bm{g}_{o_n}$.
When conflicts are detected, we adjust $\bm{g}_{o_m}$ by projecting it onto the plane that is orthogonal to the conflict direction:
%
\begin{equation}
\small
    \tilde{\bm{g}}_{o_m} = \bm{g}_{o_m} - \sum_{o_n\neq {o_m}}\frac{min(\langle \bm{g}_{o_m}, \bm{g}_{o_n} \rangle,0)}{\|\bm{g}_{o_n}\|^2} \bm{g}_{o_n},
\end{equation}
where $\min(\langle \bm{g}_{o_m}, \bm{g}_{o_n} \rangle,0)$ aims to select conflict vectors ($\langle \bm{g}_{o_m}, \bm{g}_{o_n} \rangle < 0$). Finally, we update outer loop meta-loss for user $u_i$ based on the orthogonal gradient, i.e., $\bm{\Theta}_{i} = \bm{\Theta}_{i} - \eta_2 \cdot \sum_{o_m}\tilde{\bm{g}}_{o_m}$. In summary, Algorithm~\ref{alg:booml} illustrates the whole optimization process of our BOOML.

\section{Experiments and Analysis}\label{sec:experiments}
We conduct extensive experiments on three real-world datasets to verify the efficacy of our proposed method {BOOML} by answering the following four research questions\footnote{Our code is available at \url{https://anonymous.4open.science/r/BOOML-2A75}}:
\begin{itemize}[leftmargin=*]
    \item[] \textbf{RQ1}: How does BOOML perform compared with state-of-the-art (SOTA) multi-objective recommendation approaches?
    \item[] \textbf{RQ2}: How do different components of BOOML affect its performance regarding effectiveness and efficiency? 
    \item[] \textbf{RQ3}: How does BOOML perform across different user groups? 
    \item[] \textbf{RQ4}: How do essential hyper-parameters affect the performance of our proposed BOOML?
\end{itemize}


\subsection{Experimental Setup}

\begin{table}[t]
\footnotesize
\centering
\caption{The statistics of the datasets in our study.}\label{tab:statistics}
\vspace{-0.15in}
\begin{tabular}{|c|ccccc|}
    \specialrule{.15em}{.05em}{.05em}
    & \#Users &\#Items &\#Interactions &\#Categories &Density  \\
    \specialrule{.05em}{.05em}{.05em}
    Games &13,698&42,458&160,801&471&2.7649e-4\\
    Electronic &20,247&11,589&347,393&528&1.4805e-3\\
    Movie &33,326&21,901&958,986&77&1.3139e-3\\
    \specialrule{.15em}{.05em}{.05em}     
\end{tabular}
\vspace{-0.1in}
\end{table}
\begin{table*}[!ht]
    \centering
    \setlength\tabcolsep{2.2pt}
    \renewcommand{\arraystretch}{1.1}
    \footnotesize
    \caption{Performance of all methods. The best results of BOOML are in bold; the best results of baselines are underlined; and 
    `\textit{Improvement}' indicates the relative improvements of BOOML over the strongest baseline on overall performance of multi-objectives (i.e., ResSum and HarMean). `-' denotes omitting the improvement on specific-objective metrics (i.e., NDCG, ILD, and ARP) because measuring specific-objective metrics may lead to unfair comparisons.   
    }
    \label{tab:performance}
    \vspace{-0.15in}
    \begin{tabular}{|c|ccccc|ccccc|ccccc|}
    \specialrule{.15em}{.05em}{.05em}
    \multicolumn{1}{|c|}{\cellcolor{blue!25}{$\bm{K=20}$}}     & \multicolumn{5}{c|}{Games} & \multicolumn{5}{c|}{Electronics} & \multicolumn{5}{c|}{Movies}
    \\
    
    \hline
    
    Method     & NDCG$\uparrow$ &ILD$\uparrow$ &ARP$\downarrow$ &ResSum$\uparrow$ &HarMean$\uparrow$ & NDCG$\uparrow$ &ILD$\uparrow$ &ARP$\downarrow$ &ResSum$\uparrow$ &HarMean$\uparrow$ & NDCG$\uparrow$ &ILD$\uparrow$ &ARP$\downarrow$ &ResSum$\uparrow$ &HarMean$\uparrow$
    \\
    
    \hline
 
    SMORL-MF   & 0.0420      & 0.6969      & 73.5281      & 1.2129     & 0.1099 & 0.1645     & 3.9121      & 499.9432     & 1.6454     & 0.3298 & 0.0460      & 0.5392      & 414.6145     & 1.1782     & 0.1185
    \\
    SMORL-LGCN & 0.2985     & 4.8732      & 42.1538      & 1.7968    & 0.4736 & \underline{0.4929}      & 5.2640      & 265.4003     & \underline{1.9887}     & \underline{0.5964} & 0.2734      & 7.9906      & 168.8871     & \underline{1.7745}     & 0.4510
    \\
    GFN4Rec-MF &0.0192 & 3.5574&26.6617 &1.5009&0.0545 &0.0401	&8.7562&111.5301 &1.5422	&0.1074 &0.0261	&13.9768	&155.2426	&1.5277	&0.0726
    \\
    GFN4Rec-LGCN &0.0190	&3.4943	&26.6294	&1.4989 &0.0539 &0.0385	&\underline{8.8555}	&112.0304	&1.5406	&0.1036 &0.0268	&\underline{14.1516}	&154.8179	&1.5284	&0.0744
    \\
    FairRec-MF & 0.0401 & 0.4174  & 61.5791 & 1.1470 & 0.1050  & 0.1358 & 1.0946  & 361.1347 & 1.3857 & 0.2805  & 0.0461 & 1.1019  & 384.5379 & 1.2974 & 0.1199
    \\
    FairRec-LGCN & 0.0282 & \underline{5.0225}  & 46.3841 & 1.5270 & 0.0780  & 0.1141 & 2.6632  & 332.2624 & 1.5497 & 0.2536  & 0.0173 & 7.1133  & 91.7545  & 1.5192 & 0.0493
    \\ 
    TFROM-MF  & 0.0038  & 0.3981  & 7.0588  & 1.1374 & 0.0112  & 0.0097  & 2.1665  & \underline{27.1900} & 1.4161 &0.0283   & 0.0111  & 2.5709  & \underline{56.6663}& 1.4445 & 0.0322
    \\
    TFROM-LGCN & 0.0039  & 3.7411  & \underline{6.5763}  & 1.5187 & 0.0116  & 0.0100  & 1.5123  & 27.4477 & 1.3385 & 0.0291 & 0.0120  & 7.3818  & 65.6358 & 1.5152 & 0.0348  
    \\
    DGRec & \underline{0.5441}    & 4.5535   & 32.7887    & \underline{2.0413}   & \underline{0.6226}  & 0.2960   & 3.4631    & 491.2899   & 1.7661   & 0.4682 & \underline{0.3078}    & 3.3197    & 484.9465   & 1.7734   & \underline{0.4775}
    \\
    MMoE & 0.0133 & 4.2054  & 16.2038& 1.5140 & 0.0384  & 0.0953 & 0.9850  & 246.5437 & 1.3244 & 0.2164  & 0.0338 & 0.4519  & 188.6970 & 1.1462 & 0.0903
    \\\hline  
    
    BOOML-MF & 0.0219 & \textbf{8.1789}  & 21.2125  & 1.5334 & 0.0617 & 0.0457 & \textbf{28.5732} & 123.4855 & 1.5477 & 0.1206  & 0.0319 & \textbf{61.2825} & \textbf{158.2216} & 1.5335 & 0.0874
    \\
     BOOML-LGCN & \textbf{0.6819} & 2.8868  & \textbf{15.3181}  & \textbf{2.1454} & \textbf{0.6728} & \textbf{0.7373} & 8.1628  & \textbf{49.8504}  & \textbf{2.2420} & \textbf{0.6918} & \textbf{0.4534} & 25.1284 & 424.8415 & \textbf{1.9540} & \textbf{0.5766}
     \\

     \hline  

    \textit{Improvement} &- &- & - & 5.10\% &8.06\% &- &- & - &12.74\%& 16.00\% &- &-&-&10.12\%& 20.75\%
    \\    
    \specialrule{.05em}{.05em}{.05em}
    \specialrule{.05em}{.05em}{.05em}
    \multicolumn{1}{|c|}{\cellcolor{blue!25}{$\bm{K=50}$}}  & \multicolumn{5}{c|}{Games} & \multicolumn{5}{c|}{Electronics} & \multicolumn{5}{c|}{Movies}\\\hline
    Method    & NDCG$\uparrow$ &ILD$\uparrow$ &ARP$\downarrow$ &ResSum$\uparrow$ &HarMean$\uparrow$ & NDCG$\uparrow$ &ILD$\uparrow$ &ARP$\downarrow$ &ResSum$\uparrow$ &HarMean$\uparrow$ & NDCG$\uparrow$ &ILD$\uparrow$ &ARP$\downarrow$ &ResSum$\uparrow$ &HarMean$\uparrow$\\\hline

    SMORL-MF & 0.0601      & 0.6689      & 62.9784      & 1.2253     & 0.1490 & 0.1995      & 2.7076      & 390.3489     & 1.6376     & 0.3714 & 0.0645      & 0.6121      & 362.8768     & 1.2136     & 0.1575
    \\
    SMORL-LGCN & 0.3307      & \underline{4.9965}      & 34.1166      & 1.8313     & 0.4999 & \underline{0.5041}      & 5.3842      & 199.8682     & \underline{2.0008}     & \underline{0.6020} & 0.2995       & 8.3824      & 146.2646     & 1.8010     & 0.4738 
    \\
    GFN4Rec-MF &0.0327 &3.5352 &26.5262 &1.5138 &0.0894  &0.0674 &	8.8479&	110.973&	1.5695&	0.1683 &0.0411	&13.6291	&145.5202	&1.5428&	0.1098
    \\
    GFN4Rec-LGCN &0.0330	&3.4943	&26.6295	&1.5129	&0.0901 &0.0647	&\underline{8.9393}	&111.5699	&1.5668	&0.1626 &0.0414	&\underline{13.7655}	&145.1359	&1.5431	&0.1105
    \\
    FairRec-MF & 0.0625 & 0.4497  & 68.1659 & 1.1767 & 0.1529  & 0.1931 & 1.1984  & 404.3716 & 1.4620 & 0.3539  & 0.0672 & 1.3663  & 401.0979 & 1.3646 & 0.1654
    \\
    FairRec-LGCN & 0.0420 & 4.6429  & 44.3422 & 1.5381 & 0.1120  & 0.1553 & 2.7858  & 312.7333 & 1.5980 & 0.3159  & 0.0237 & 5.8455  & 72.8516  & 1.5242 & 0.0664 
    \\
    TFROM-MF & 0.0070  & 0.4226  & 6.7909  & 1.1479 & 0.0205  & 0.0172  & 2.3471  & \underline{28.4492} & 1.4387 & 0.0490 & 0.0186  & 2.5963  & \underline{60.1280} & 1.4534 & 0.0528 
    \\
    TFROM-LGCN  & 0.0070  & 3.7343  & \underline{6.7461}  & 1.5207 & 0.0206  & 0.0183  & 1.6188  & 28.8897 & 1.3616 & 0.0519  & 0.0192  & 7.4751  & 60.8851 & 1.5227 & 0.0545 
    \\ 
    {DGRec} & \underline{0.5669}    & 4.5793    & 30.1526    & \underline{2.0650}   & \underline{0.6327} & 0.3357    & 3.4342    & 407.3286   & 1.8051   & 0.4993 & \underline{0.3459}    & 3.3442     & 404.1047   & \underline{1.8124}   & \underline{0.5064}
    \\
    MMoE & 0.0202 & 4.9100  & 14.5558 & 1.5300 & 0.0572  & 0.1289 & 0.9770  & 209.3083 & 1.3566 & 0.2696   & 0.0481 & 0.4606  & 174.1520 & 1.1627 & 0.1229
    \\\hline  
    BOOML-MF & 0.0321 & \textbf{7.7124}  & 17.1521 & 1.5462 & 0.0880  & 0.0722 & \textbf{26.1038} & 115.2841 & 1.5744  & 0.1781 & 0.0453 & \textbf{57.8995} & \textbf{130.8545} & 1.5472 & 0.1197 
    \\ 
    BOOML-LGCN & \textbf{0.6636} & 2.9633  & \textbf{14.9774} & \textbf{2.1312} & \textbf{0.6676} & \textbf{0.6901} & 8.3259  & \textbf{47.2916}  & \textbf{2.1951} & \textbf{0.6774} & \textbf{0.4515} & 26.5220 & 373.6074 & \textbf{1.9522} & \textbf{0.5756}
    \\

    \hline
    \textit{Improvement}&- &-&-&3.21\%&5.52\% &-&-&-&9.71\%&12.52\%&-&-&-&7.71\%&13.67\% \\
    \specialrule{.15em}{.05em}{.05em}
    \end{tabular}
    \vspace{-0.1in}
\end{table*}

\subsubsection{Datasets} We adopt three real-world datasets with varying domains, sizes, and sparsity levels collected from Amazon.com~\cite{ni2019justifying}, including Games, Electronics, and Movies. The datasets contain users' ratings on the scale of $[1,5]$ stepped by 1 towards products in the three domains. Following~\cite{sun2022daisyrec}, we convert the interactions with ratings no less than 4 as positive feedback; otherwise negative feedback. Besides, we filter out users and items with less than 10 interactions. Table~\ref{tab:statistics} shows the statistics of the three datasets after pre-processsing. Finally, each dataset is chronologically split into training, validation, and test sets in a 6:2:2 ratio.

\subsubsection{Evaluation Metrics} As introduced in Section~\ref{sec:objectives}, we adopt the widely-used NDCG@K, ILD@K, and ARP@K to evaluate the performance of accuracy, diversity, and fairness, respectively.  
Additionally, we also adopt the Rescaled Sum (ResSum@K) and Harmonic Mean (HarMean@K) as defined in Section~\ref{sec:BO}, to evaluate the comprehensive performance of all methods. 
In particular, larger NDCG and ILD, ResSum, and HarMean values indicate better performance, whereas smaller ARP values suggest fairer recommendations.
We set $K=\{20, 50\}$ in our study empirically.

\subsubsection{Baselines} We compare with six SOTA multi-objective RSs at Levels 3-4. Specifically, 
\textbf{DGRec}~\cite{yang2023dgrec} is a diversifying GNN-based RS at Level 3, which directly improves the embedding generation procedure for diversified recommendations. 
\textbf{SMORL}~\cite{stamenkovic2022choosing} is a reinforcement learning based RS at Level 3, which augments recommenders with additional neural layers to optimize three objectives: accuracy, diversity, and novelty.  
\textbf{GFN4Rec}~\cite{liu2023generative} is a generative RS at Level 3, which aims to learn a policy that can generate sufficiently diverse item lists for users while maintaining high recommendation quality. 
\textbf{FairRec}~\cite{patro2020fairrec} is a scalable and adaptable RS at Level 3, which ensures uniform fairness for products by setting the minimum exposure, and fairness for users using a greedy strategy. 
\textbf{TFROM}~\cite{wu2021tfrom} is a post-processing RS at Level 3, which designs heuristic algorithms to ensure two-sided fairness at the cost of reduced recommendation quality.  
\textbf{MMoE}~\cite{ma2018modeling} is a generic multi-objective RS at Level 4 that can optimize accuracy, diversity, and fairness using Mixture-of-Experts, explicitly learning to model task relationships.

{As our BOOML and most baseline methods (i.e., SMORL, GFN4Rec, FairRec, and TFROM) need to be built on existing user and item encoders, we further choose two representative encoders to verify their generality, including non-graph-based encoder \textbf{MF}~\cite{sun2022daisyrec} and graph-based encoder \textbf{LGCN}~\cite{he2020lightgcn}.}

\begin{table*}[!ht]
\centering
\fontsize{7.5}{7.5}\selectfont
\setlength\tabcolsep{1.5pt} 
\renewcommand{\arraystretch}{1.3}
\caption{Performance of different variants of our BOOML. The best performance for each metric is highlighted in bold.}\label{tab:ablation-study}
\vspace{-0.15in}
\begin{tabular}{|c|c|ccccc|ccccc|ccccc|}
\specialrule{.15em}{.05em}{.05em}
\multicolumn{2}{|c|}{\cellcolor{blue!25}{$K=50$}} 
& \multicolumn{5}{c|}{Games}                         
& \multicolumn{5}{c|}{Electronics}                    
& \multicolumn{5}{c|}{Movies}                        \\ \hline
& Variant              
& \multicolumn{1}{c|}{NDCG $\uparrow$} 
& \multicolumn{1}{c|}{ILD $\uparrow$}  
& \multicolumn{1}{c|}{ARP $\downarrow$}
& \multicolumn{1}{c|}{ResSum $\uparrow$}
& \multicolumn{1}{c|}{Epoch $\downarrow$}
& \multicolumn{1}{c|}{NDCG $\uparrow$} 
& \multicolumn{1}{c|}{ILD $\uparrow$}  
& \multicolumn{1}{c|}{ARP $\downarrow$}
& \multicolumn{1}{c|}{ResSum $\uparrow$} 
& \multicolumn{1}{c|}{Epoch $\downarrow$}
& \multicolumn{1}{c|}{NDCG $\uparrow$} 
& \multicolumn{1}{c|}{ILD $\uparrow$}  
& \multicolumn{1}{c|}{ARP $\downarrow$}
& \multicolumn{1}{c|}{ResSum $\uparrow$} 
& \multicolumn{1}{c|}{Epoch $\downarrow$} \\ \hline
\specialrule{.05em}{.05em}{.05em}
\multirow{4}{*}{\rotatebox[origin=c]{90}{MF}}       
& SGD  & 0.0627      & 0.8099      & 67.0925     & 1.2585       & 50
& 0.2060      & 0.8869      & 429.2563     & 1.4148       & 40
& 0.0798      & 1.2949      & 412.3657     & 1.3654       & 50 \\ 
& BO   & \textbf{0.0631}      & 0.5985      & 66.3256     & 1.2122       & 50
& 0.2072      & 1.3842      & 427.9066     & 1.5074       & 40
& \textbf{0.0811}      & 1.6076      & 409.4536     & 1.4148       & 50 \\ 
& BOML  & 0.0300      & 7.0660      & 24.1939     & 1.5395       & 5
& 0.0569      & 16.6171     & \textbf{91.7634}      & 1.5596       & 5
& 0.0472      & 39.5236     & 153.1433     & \textbf{1.5488}    & 5 \\ 
& BOOML & 0.0321      & \textbf{7.7124}  & \textbf{17.1521} & \textbf{1.5462}  & 5
& \textbf{0.0722} & \textbf{26.1038} & 115.2841     & \textbf{1.5744}       & 5
& 0.0453      & \textbf{57.8995} & \textbf{130.8545} & 1.5472       & 5 \\ 
\specialrule{.05em}{.05em}{.05em}
\multirow{4}{*}{\rotatebox[origin=c]{90}{LGCN}}
& SGD  & 0.3208      & 2.0880      & 36.3096     & 1.7174       & 75
& 0.2911      & 2.6535      & 332.1371     & 1.7261       & 60
& 0.1392      & 4.1039      & 103.0174     & 1.6254       & 80 \\ 
& BO   & 0.3355      & 2.6069      & 30.7014     & 1.7749       & 75
& 0.3157      & 3.1066      & 309.8266     & 1.7737       & 60
& 0.2028      & 4.8385      & \textbf{66.5464}      & 1.6987       & 80 \\ 
& BOML  & \textbf{0.7030}      & \textbf{3.0340}      & \textbf{6.8549}      & \textbf{2.1935}       & 1
& \textbf{0.7162}      & 7.4462      & 112.3567     & \textbf{2.2178}       & 1
& \textbf{0.5265}      & 18.5016     & 276.1999     & \textbf{2.0274}       & 1 \\ 
& BOOML & 0.6636      & 2.9633      & 14.9774     & 2.1312       & 1
& 0.6901      & \textbf{8.3259}  & \textbf{47.2916}  & 2.1951       & 1
& 0.4515      & \textbf{26.5220} & 373.6074     & 1.9522       & 1 \\ 
\specialrule{.15em}{.05em}{.05em}
\end{tabular}
\vspace{-0.1in}
\end{table*}

\subsubsection{Implementation Details}
We empirically find out the optimal settings for essential hyper-parameters of each method according to the performance on the validation set. For all encoders, the batch size is set as 1024 and the embedding size is set as 64 for fair comparison. The learning rate is searched in $\{1e-1, 1e-2, 1e-3\}$, and set as 1e-3. The optimizer is searched from AdamW and SGD, where the best option is SGD for MF, while AdamW for LGCN. For the number of layers in LGCN, we search in scale $\{2,3,4\}$ and set as 2 because it shows the best performance. 

Regarding the multi-objective baselines, their hyper-parameters are searched and set as follows. 
For SMORL, the discount factor $\gamma=0.9$; the objective-balancing weight vector $\bm{w}=(1,1,1)$; the weight $\alpha$ to control the influence of SMORL is searched in $\{0.5, 1, 1.5, 2\}$ and set as 2. 
For GFN4Rec, as suggested by the paper, we set $b_z=1$; $b_r$ and $b_f$ are respectively searched in $\{0.1, 0.3, 1, 1.5\}$ and $\{0.1, 0.5, 1, 1.5, 2\}$; and the optimal settings are $b_r = 1.5$ and $b_f=1$.
For FairRec, the importance of fairness objective ($\alpha$) is searched in $[0.1, 0.9]$ stepped by 0.2, and the best option is 0.5. 
For DGRec, the learning rate is searched in $\{1e-1,1e-2,1e-3\}$ and set as $1e-1$; the number of GNN layers is searched in $[1,2,3]$ and set as 2; and the weight to control popular categories ($\beta$) is searched in $[0.9, 0.95]$ stepped by 0.01 and set as 0.93.  
For MMoE, the dropout rate is searched in $[0.1, 0.5]$ stepped by 0.1, and set as 0.2; the number of experts is searched in $[2,3,4,5]$ and set as 4; the number of layers is 2 and the hidden units per expert are $\{64,32\}$.
For our BOOML, the number of initial points is 10; the trials of BO is 50; the function $g(\cdot)$ adopts rescaled sum; $\kappa=0$ for simplicity; the inner learning rate and outer learning rate of meta-learning are searched in $\{1e-1,1e-2,1e-3\}$ and both set as $1e-2$; the meta-learning epoch is searched in $\{1,2,3,4,5\}$, and the optimal option is 5 for MF and  1 for LGCN; $W$ is searched in $\{2, 3, 4, 5\}$ and set as 3; and $\bm{\lambda}$ and $\bm{\beta}$ are searched in the range of $[0.01, 10]$.

\subsection{Results and Analysis} 

\subsubsection{Comparative Results (RQ1)} Table~\ref{tab:performance} presents the performance of all methods. Several major observations are noted. 

%
\begin{itemize}[leftmargin=0.3cm]
    \item[-] \textbf{First}, across all encoders, our BOOML demonstrates a positive \textit{improvement} on ResSum and HarMean in all cases compared to baseline methods. This highlights BOOML's superiority to balance multi-objective performance by leveraging orthogonal meta-learning to alleviate conflicts among different objectives.
    \item[-] \textbf{Second}, BOOML achieves better performance on accuracy, measured by NDCG, across all cases. However, its performance on diversity and fairness, measured by ILD and ARP, is worse than the best-performing baselines. This is attributed to some baselines focusing only on specific objectives while largely sacrificing other objectives. For example, the fairness-oriented methods TFROM-MF and TFROM-LGCN perform exceptionally well in ARP but significantly undermine both accuracy and diversity. Particularly, it consistently produces the worst NDCG in all cases compared with other methods.
    The diversity-oriented methods GFN4Rec-MF and GFN4Rec-LGCN perform well in ILD but compromise both accuracy and fairness.
    \item[-] \textbf{Third}, diversity-oriented baselines (e.g., SMORL and GFN4Rec) generally outperform fairness-oriented baselines (e.g., FairRec and TFROM) in terms of ILD across most cases. Conversely, fairness-oriented baselines defeat diversity-oriented ones regarding ARP. Additionally, the diversity-oriented method DGRec consistently surpasses the generic multi-objective baseline MMoE in both NDCG and ILD but performs worse on ARP, but DGRec achieves better overall performance than MMoE, suggesting its stronger ability to balance multiple objectives.
    \item[-] \textbf{Lastly}, different baselines show varying sensitivity to encoders across all metrics. For example, in the aspect of accuracy, BOOML and SMORL are particularly sensitive to encoders and achieve the best NDCG performance when using LGCN as the encoder. Furthermore, all baselines show sensitivity to encoders in ILD and ARP except for TFROM which remains largely insensitive to encoders for ARP.  These results underscore the importance of selecting the most suitable encoder for different multi-objective baselines to achieve optimal performance.  Beyond BOOML and SMORL, we also observe that the diversity-oriented DGRec, built on GNN, achieves relatively strong performance on NDCG, highlighting the potential of GCN/GNN structures in enhancing the accuracy of multi-objective optimization.
\end{itemize} 

\begin{table*}[t]
\fontsize{8}{7}\selectfont
\renewcommand{\arraystretch}{1.1}
\caption{The learned weights and corresponding performance for different groups across various metrics.}\label{tab:weights_across_group}
\vspace{-0.15in}
\begin{tabular}{|l|l|l|l|l|l|l|l|l|l|l|}
\specialrule{.15em}{.05em}{.05em}
\multicolumn{2}{|c|}{Encoder=MF}  
&\multicolumn{2}{c|}{Learned Weights}       
&\multicolumn{3}{c|}{Normalized Weights}
&\multicolumn{4}{c|}{Metrics ($K=20$)}   \\ \hline
Dataset   & Group & Diversity ($\lambda$) & Fairness ($\beta$)  & Accuracy    & Diversity & Fairness & NDCG$\uparrow$    & ILD$\uparrow$    & ARP$\downarrow$       & ResSum$\uparrow$ \\ 
\specialrule{.05em}{.05em}{.05em}
\multirow{3}{*}{Games}       
& $\mathcal{G}_1$ & 0.9724 & 4.7997 & 0.1477 & \cellcolor{blue!25}0.1436 & \cellcolor{green!25}0.7087 & 0.0077 & \cellcolor{blue!25}8.0829 & \cellcolor{green!25}16.4183 & 1.5226 \\ \cline{2-11}
& $\mathcal{G}_2$ & 0.0108 & 2.8620 & \cellcolor{red!25}0.2582 & 0.0028 & \cellcolor{green!25}0.7390 & \cellcolor{red!25}0.0086 & 8.0733 & \cellcolor{green!25}16.7074 & 1.5232 \\ \cline{2-11}
& $\mathcal{G}_3$ & 0.0592 & 0.6837 & \cellcolor{red!25}0.5738 & \cellcolor{blue!25}0.0340 & 0.3923 & \cellcolor{red!25}0.0493 & \cellcolor{blue!25}8.3839 & 30.4972 & 1.5573 \\ 
\specialrule{.05em}{.05em}{.05em}
\multirow{3}{*}{Electronics}    
& $\mathcal{G}_1$ & 5.3986 & 2.3176 & \cellcolor{red!25}0.1147 & 0.6194 & \cellcolor{green!25}0.2659 & \cellcolor{red!25}0.0471 & 28.4640 & \cellcolor{green!25}123.0598 & 1.5491 \\ \cline{2-11}
& $\mathcal{G}_2$ & 9.1716 & 3.3809 & 0.0738 & \cellcolor{blue!25}0.6767 & \cellcolor{green!25}0.2495 & 0.0318 & \cellcolor{blue!25}28.5024 & \cellcolor{green!25}122.7765 & 1.5338 \\ \cline{2-11}
& $\mathcal{G}_3$ & 9.9658 & 0.3105 & \cellcolor{red!25}0.0887 & \cellcolor{blue!25}0.8838 & 0.0275 & \cellcolor{red!25}0.0788 & \cellcolor{blue!25}28.8683 & 125.7153 & 1.5808 \\ 
\specialrule{.15em}{.05em}{.05em}
\end{tabular}
\vspace{-0.1in}
\end{table*}

\begin{figure*}[t]
    \centering
    \subfigure[$E_{ml}$ on MF]{
        \includegraphics[width=0.22\linewidth]{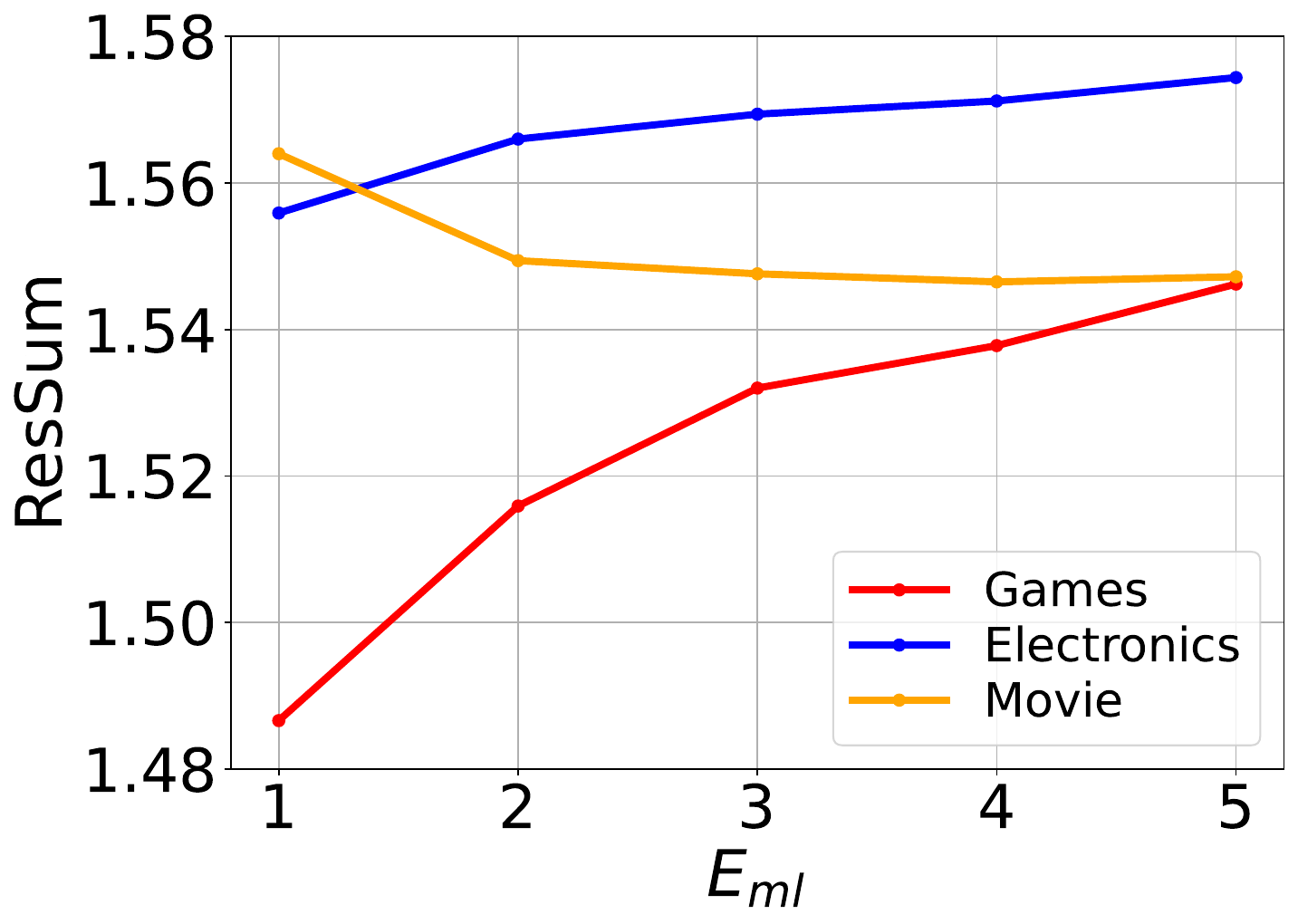}
    }
    \subfigure[$E_{ml}$ on LGCN]{
        \includegraphics[width=0.22\linewidth]{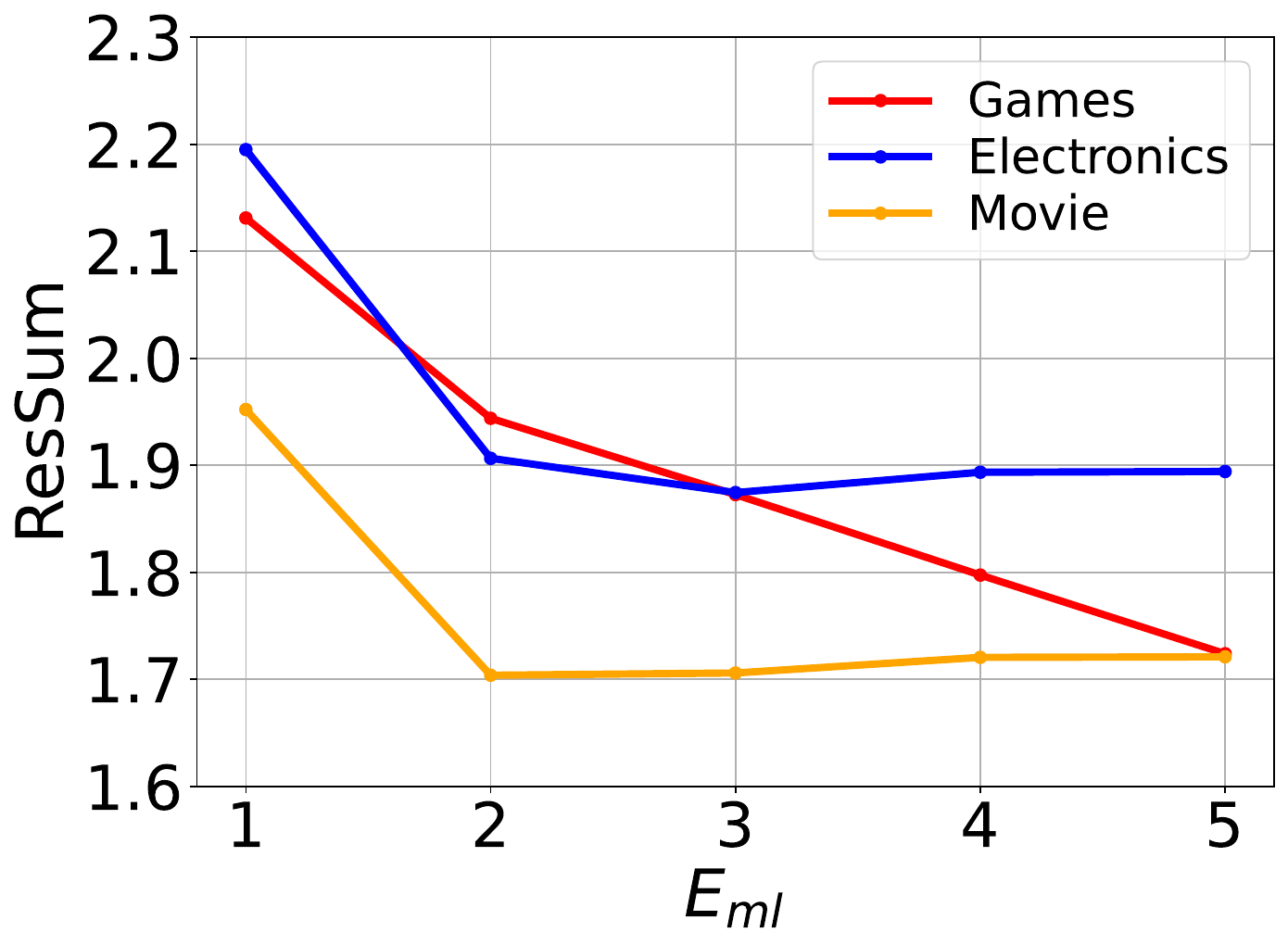}
    }
    \subfigure[$W$ on MF]{
        \includegraphics[width=0.22\linewidth]{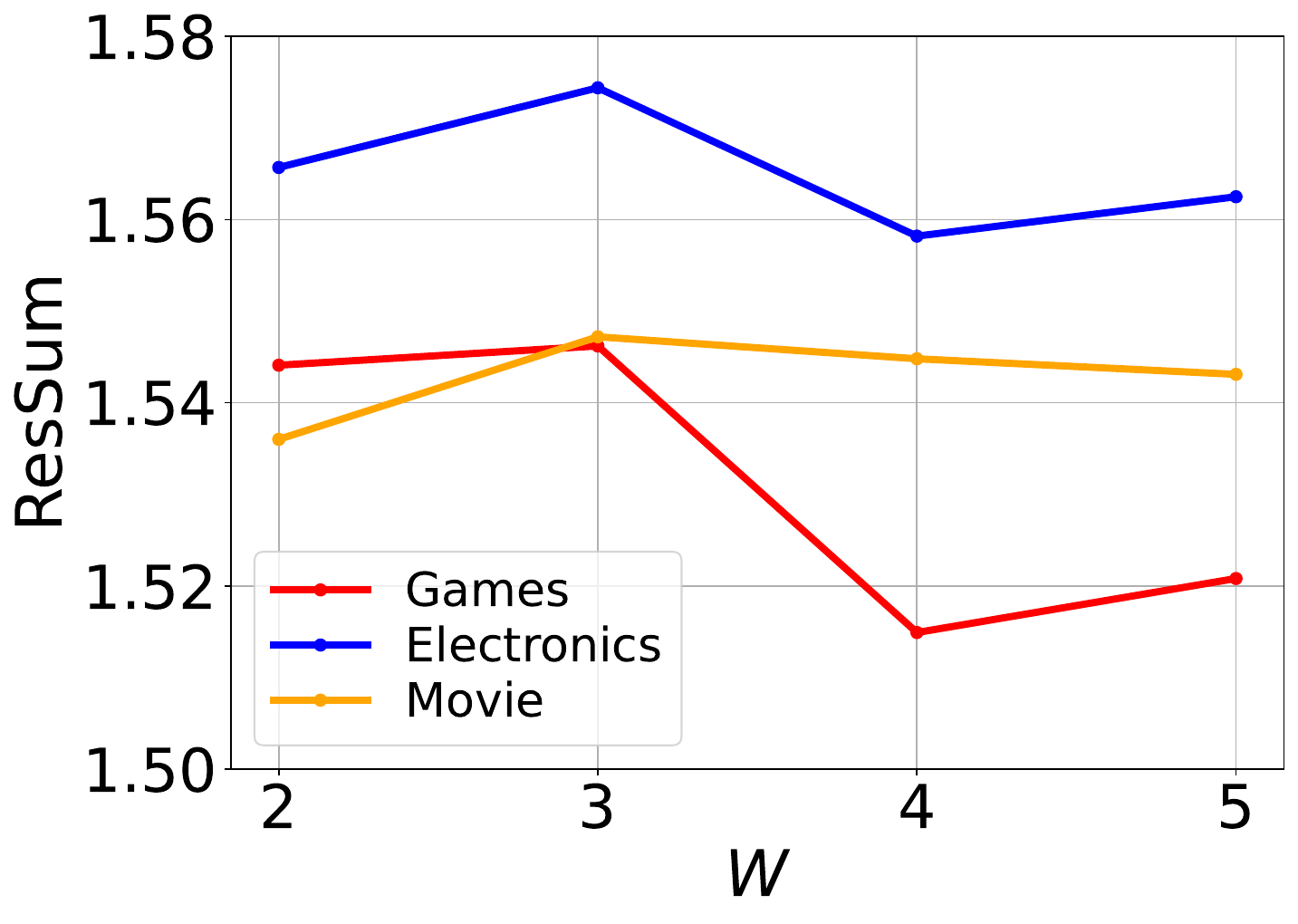}
    }
    \subfigure[$W$ on LGCN]{
        \includegraphics[width=0.22\linewidth]{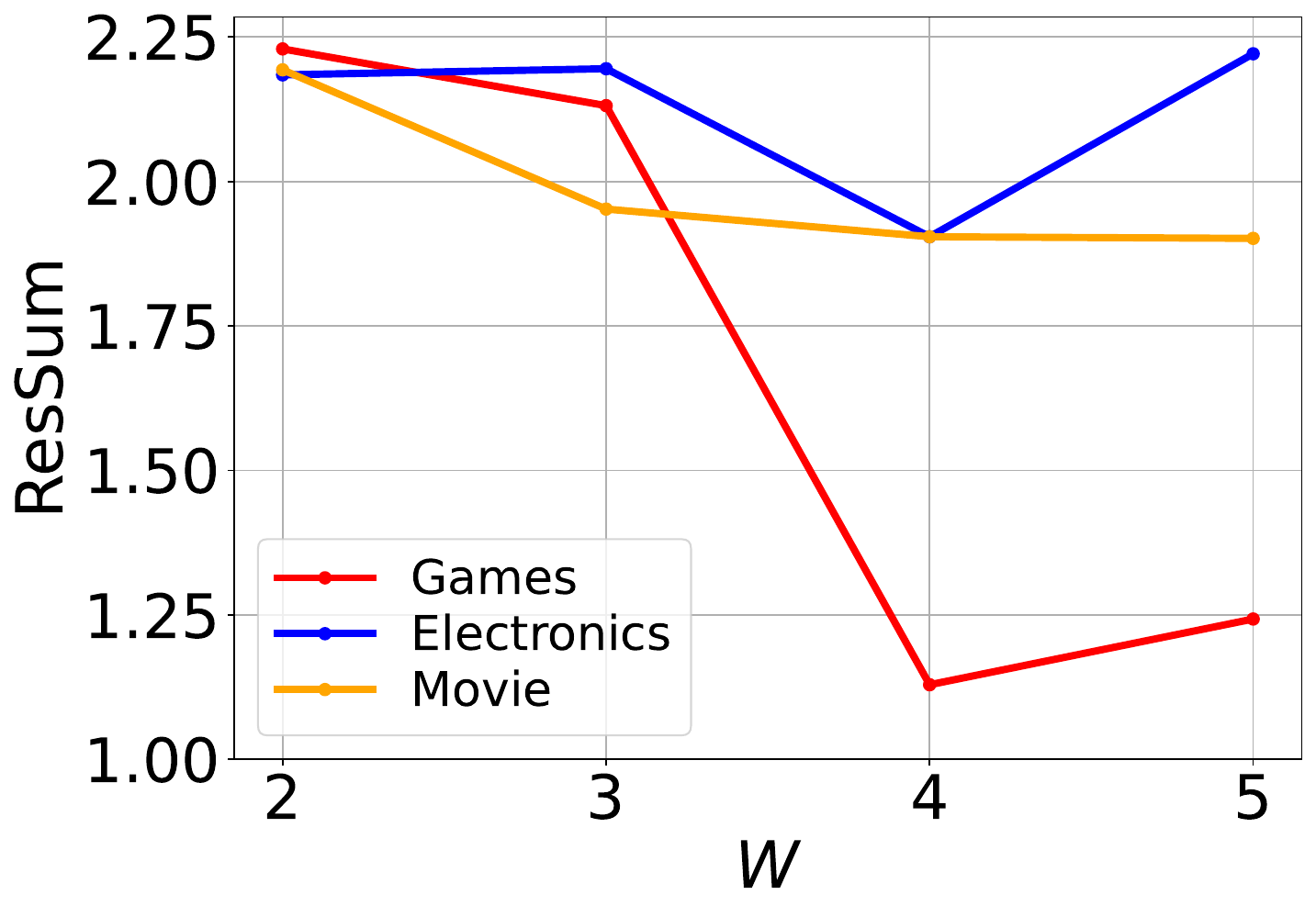}
    }
    \vspace{-0.15in}
    \caption{The impacts of meta-learning epochs $E_{ml}$ and the number of user groups $W$.}
    \label{fig:hyper-parameter}
    \vspace{-0.1in}
\end{figure*}

\subsubsection{Ablation Study (RQ2)} To examine the efficacy of different components of our BOOML, we compare it with different variants. Specifically, (1) \textbf{SGD} directly uses SGD with constant weights to optimize the recommender, i.e., $\lambda_w = \beta_w = 1.0$ for all user groups;
(2) \textbf{BO} adopts vanilla Bayesian optimization to search the optimal weights of different objectives for each group; (3) \textbf{BOML} adopts meta-learning in the BO process to search the optimal weights of different objectives by considering the correlations among different user groups; and (4) \textbf{BOOML} is our proposed method which exploits orthogonal meta-learning in the BO process by considering the correlations of different groups and potential conflicts among various objectives.  Table~\ref{tab:ablation-study} shows the performance across different evaluation metrics and training epochs of all variants with the two encoders on all datasets. Four key findings can be identified. 

\begin{itemize}[leftmargin=0.3cm]
\item[-] \textbf{First}, {BO} generally outperforms {SGD}, especially on the diversity and fairness metrics (i.e., ILD and ARP), which verifies the necessity and effectiveness of using BO to search for the optimal weights for better multi-objective performance.
\item[-] \textbf{Second}, compared with BO, BOML generally delivers superior performance with LGCN as the encoder. However, BOML exhibits lower performance on NDCG when using MF as encoders but gains significant improvements in diversity and fairness metrics. For example, BOML-MF results in a 52\% decrease on Games dataset in NDCG, it achieves a 108\% improvement in ILD and a 63.52\% reduction in ARP, ultimately leading to a 27\% increase in ResSum. These results, on one hand, highlight the effectiveness of meta-learning in improving recommendation performance by capturing correlations among different user groups; on the other hand, they reveal that potential conflicts among objectives may constrain the overall effectiveness of meta-learning. 
\item[-] \textbf{Third}, BOOML-MF generally defeats BOML-MF across most metrics, resulting in overall performance improvements. This highlights BOOML's capability to mitigate potential conflicts among different objectives. However, BOOML-LGCN underperforms BOML-LGCN in most cases. This may be attributed to BOOML's reduced conflict mitigation capability during the layer-wise propagation process with the GCN encoder. 
\item[-] \textbf{Lastly}, BOML and BOOML achieve better performance than SGD and BO in most cases with significantly fewer training epochs. For instance, the training epochs for SGD and BO are in the range of $[40,80]$, while BOML and BOOML only require 5 meta-learning epochs. This verifies that the meta-optimization and orthogonal gradient descent greatly enhance the training efficiency and effectiveness.   
\end{itemize}
\subsubsection{Performance Across Different Groups (RQ3)}
Table~\ref{tab:weights_across_group} shows the learned weights for different objectives and the corresponding performance across different metrics of our BOOML-MF on Games and Electronics, respectively. In the table, the `Learned Weights' means the original weights learned by our BOOML for different objectives. For ease of analysis, we also calculate the `Normalized Weights'; we highlight the higher weights for different objectives (e.g., Accuracy) across different groups; and the corresponding metrics (e.g., NDCG) with better results are highlighted in the same color. For instance, on Games, across the three groups, $\mathcal{G}_2$ and $\mathcal{G}_3$ have higher weights on Accuracy than $\mathcal{G}_1$, so they are highlighted in red; and $\mathcal{G}_2$ and $\mathcal{G}_3$ achieve the best NDCG values, so they are also highlighted in red. Similarly, we highlight the higher weights on `Diversity' and `Fairness' in blue and green, respectively. 

From the results, two observations can be noted. \textbf{First}, \textit{on all datasets, across different groups, the objectives with higher weights gain better results on the corresponding metrics}. For instance, on Games, $\mathcal{G}_2$ and $\mathcal{G}_3$ have higher weights on Accuracy (0.2582 and 0.5738) than $\mathcal{G}_1$ (0.1477), thus they gain better results on NDCG (0.0086 and 0.0493) than that of $\mathcal{G}_1$ (0.0077); while $\mathcal{G}_1$ and $\mathcal{G}_2$ possess higher weights on Fairness (0.7087 and 0.7390) than $\mathcal{G}_3$ (0.3923), so they obtain better results on ARP (16.4183 and 16.7074) than that of $\mathcal{G}_3$ (30.4972). This helps verify that our BOOML can better uncover the uncertain relationships between the weights and performance of different objectives. 
\textbf{Second}, \textit{different groups inherently prioritize distinct multiple objectives}. For instance, based on the learned weights and performance of different objectives across each group, on Games, we observe that users in $\mathcal{G}_1$ place greater emphasis on both Diversity and Fairness, $\mathcal{G}_2$ prioritize Accuracy and Fairness, while $\mathcal{G}_3$ focus more on Accuracy and Diversity.

\subsubsection{Hyper-parameter Analysis (RQ4)}
We now examine how essential hyper-parameters affect the performance of our BOOML-MF and BOOML-LGCN, including the number of meta-learning epochs ($E_{ml}$) and the number of user groups ($W$). Figure~\ref{fig:hyper-parameter} depicts the results, where we vary $E_{ml}$ in the range of $[1,5]$ and $W$ in the range of {$[2,5]$}, with both stepped by 1.  
For $E_{ml}$, the performance generally improves as $E_{ml}$ increases and then keeps relatively stable with BOOML-MF. In contrast, the performance of BOOML-LGCN consistently declines as $E_{ml}$ increases. As explained, this may be attributed to BOOML's diminished conflict mitigation capability during the layer-wise propagation process in GCN. Thus, we set $E_{ml} = 5$ for MF and $E_{ml} = 1$ for LGCN.
Regarding $W$, in most cases, the performance increases initially, reaches a peak, and then declines. To ensure consistency and simplicity, we suggest to set $W = 3$ in real-world application.

\section{Conclusion and Future Work}
In this paper, we introduce a novel framework defining five levels of autonomy for RSs based on their ability to independently determine recommendation objectives. Accordingly, we propose an orthogonal meta-learning boosted Bayesian optimization approach to automatically identify and optimize uncertain multi-objectives (i.e., accuracy, diversity and fairness) based on individual user needs. 
Specifically, it leverages BO to explore the search space and quantify uncertainties between the weights and overall objectives, where the orthogonal meta-learning paradigm significantly improves optimization efficiency and effectiveness through collaborative information sharing and objective conflict reduction. Experimental results demonstrate that our approach can better optimize uncertain multi-objectives for individual users compared with SOTAs, taking a significant step toward more ethical and user-centric RSs.
For future works, we plan to (1) incorporate temporal dynamics to adapt to evolving user preferences and objectives over time and (2) expand the framework to address additional ethical concerns, e.g., transparency and privacy, enhancing the societal impact of RSs.

\bibliographystyle{ACM-Reference-Format}
\bibliography{reference}

\end{document}